# Cross-institution text mining to uncover clinical associations: a case study relating social factors and code status in intensive care medicine

Madhumita Sushil[1], Atul J. Butte[1], Ewoud Schuit[2], Maarten van Smeden[2], Artuur M. Leeuwenberg[2,*]

1 Bakar Computational Health Sciences Institute, University of California, San Francisco, San Francisco, USA

2 Julius Center for Health Sciences and Primary Care, University Medical Center Utrecht, Utrecht University, The Netherlands

**\*** Corresponding author: Artuur M. Leeuwenberg, Email: A.M.Leeuwenberg-15@umcutrecht.nl, Str. 6.131, Universiteitsweg 100, 3508 GA Utrecht, The Netherlands.

**Abstract**

**Objective:** Text mining of clinical notes embedded in electronic medical records is increasingly used to extract patient characteristics otherwise not or only partly available, to assess their association with relevant health outcomes. As manual data labeling needed to develop text mining models is resource intensive, we investigated whether off-the-shelf text mining models developed at external institutions, together with limited within-institution labeled data, could be used to reliably extract study variables to conduct association studies.

**Materials and Methods:** We developed multiple text mining models on different combinations of within-institution and external-institution data to extract social factors from discharge reports of intensive care patients. Subsequently, we assessed the associations between social factors and having a do-not-resuscitate/intubate code.

**Results:** Important differences were found between associations based on manually labeled data compared to text-mined social factors in three out of five cases. Adopting external-institution text mining models using manually labeled within-institution data resulted in models with higher F1-scores, but not in meaningfully different associations.

**Discussion:** While text mining facilitated scaling analyses to larger samples leading to discovering a larger number of associations, the estimates may be unreliable. Confirmation is needed with better text mining models, ideally on a larger manually labeled dataset.

**Conclusion:** The currently used text mining models were not sufficiently accurate to be used reliably in an association study. Model adaptation using within-institution data did not improve the estimates. Further research is needed to set conditions for reliable use of text mining in medical research.

## Introduction

Electronic health records (EHR) can be a valuable resource for medical studies measuring multivariable associations between patient characteristics and subsequent health outcomes [1–4], commonly measured in epidemiological studies [5–8]. A large part of the information is expressed in free text clinical notes and is largely unused. Because manual extraction of study variables from free text is time-consuming and requires domain expertise [9–12], text mining is increasingly being used to extract this information [13–15], to be consequently used in medical studies [16–20]. Due to differences in use of EHR systems, terminology, and writing style, off the shelf text mining models may not generalize well from one institution to the other. Models developed at an external center may require adaptation to the local target setting (e.g., via model fine-tuning) [21]. As a result of a lack of generalizability of the text mining model, consecutive medical study results may be biased. However, not all errors in measurement necessarily lead to consequent bias in medical study results [22,23]. In a case study associating social factors to code status — the type of urgent life-saving measures indicated if a patient's heart or breathing were to stop — in intensive care unit patients, we set out to assess if a text mining model developed on data from a one institution can be used to conduct an association study in a different institution. We investigate whether having a small amount of locally manually labeled data can help to more reliably estimate the target association, either via fine tuning the external text mining model, or by retraining a new model from scratch.

## Background and Significance

Sociodemographic factors, also referred to as social determinants of health (SDoH) factors, encompass several non-clinical factors — including living, working, and environmental conditions — that can have a significant impact on an individual's health [24]. Z codes represent these factors in the ICD-10 hierarchy, but they are populated for only nearly 2% of the patients in EHRs [25–27]. More frequently, SDoH factors are discussed in clinical notes as free-text rhetoric [28]. Several rule- and machine learning-based natural language processing (NLP) systems have been developed to extract SDoH information from clinical notes [12,28–36]. Recently, an N2C2 shared task (track 2, 2022) was organized for the extraction of information about substance use (alcohol/drugs/tobacco), employment, and living status from clinical notes[1]. The shared task provided manually labeled data from two different institutions to address three different methodological challenges: (a) developing methods for SDoH extraction using within-institution data, (b) developing methods that generalize from one institution to another without any access to the data from the external institution, (c) developing techniques to transfer SDoH extraction models from one institution to another, while having access to data from both institutions. Significant drops in extraction performance were seen in setting (b) of the shared task compared to settings (a) and (c), i.e., systems developed without using manually labeled within-institution data were significantly worse than the systems developed on within-institution data. As manually labeling data is time-consuming and requires domain expertise, the current study aimed to assess the impact of having within-institution manually labeled data on downstream results of medical studies relating SDoH factors to patient health outcomes.

Text-extracted SDoH factors have already been associated with diverse health outcomes like mental health impact, unplanned hospital admission/readmission, and development of chronic and non-chronic diseases including cardiovascular diseases, respiratory diseases, and cancer [37]. Less frequently documented conditions like employment status are also less frequently included in such studies, and vice versa. In our study, we included both frequently documented SDoH, such as substance use information, as well as less frequently documented SDoH factors, such as employment and living status.

---

[1] https://n2c2.dbmi.hms.harvard.edu/2022-track-2

Do-not-resuscitate (DNR) and do-not-intubate (DNI) code status during hospital stay are important predictors for several clinical outcomes like length of hospital stay, survival, and disease progression [38–42]. Although some previous studies have examined the impact of SDoH factors on code status [43–46], these relations are still not clearly understood. This is particularly true for factors that are frequently absent from structured data sources, such as employment status, living status, and information about substance use. Significant associations between such factors and code status could indicate an impact on health outcomes for specific socioeconomic patient cohorts and merits further investigation. In our pilot study, we explore whether mentions of being employed, living together with others, and having no history of substance use in discharge reports is associated with presence of DNR and DNI code status.

**Methods**

*Study data*

Electronic health record data were used from the MIMIC-III database [47]. This database contains both the structured data as well as the clinical notes for 58,976 critical care unit admissions for 46,520 patients between 2001 and 2012 at the Beth Israel Deaconess Medical Center. In this study, patients between 18-89[2] years were included, for whom the discharge summaries were linkable to the structured patient characteristics. For a subset of the admissions, the social history sections of the discharge summaries were manually labeled with SDoH factors by Lybarger et al., 2021 [12], which are made available via the N2C2 shared task[3]. These include employment status, living status, and substance use (tobacco, alcohol, and drugs). The complete data was split into three subsets: (1) a large unlabeled set of admissions for which SDoH are extracted automatically from social history mentioned in discharge reports via the studied text mining models, (2) a smaller manually labeled subset of admissions which are used to evaluate the text mining models, but also to compare the identified associations in the manually labeled and the extracted SDoH data, and (3) a small manually labeled set of 500 admissions used to obtain a local text mining model, either by fine-tuning an external model, or by training a new model from scratch only on within-institution data. The flow diagram for patient selection is presented in **Figure 1**.

---

[2] In MIMIC-III, all patients with age > 89 were given age 300 as part of de-identification.
[3] https://n2c2.dbmi.hms.harvard.edu/2022-track-2

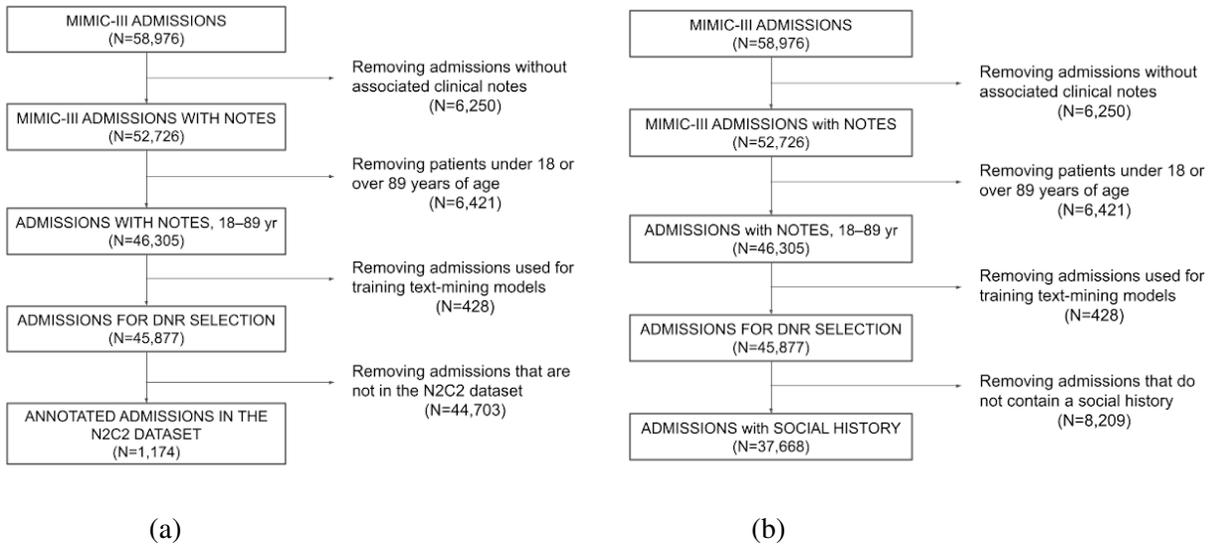

**Figure 1**: Flow diagrams reflecting the patient selection criteria for patients: **a)** in the manually labeled corpus, as well as **b)** from the complete MIMIC-III database.

*Overall study design*

The overall study design is schematically depicted in **Figure 2**. Three different approaches of using text mining for SDoH extraction are compared: (1) using a set of within-institution manually labeled data to train a new model, (2a) using a model developed at an external institution, or (2b) using an external model, but first adopting it it to the local institution (via fine-tuning) with a small set of locally available manually labeled data. Then, we assess the association between different SDoH and DNR/DNI status, based on the SDoH extracted by each respective strategy. The found associations are compared with each other to assess the impact of the strategy on consequent study results. We perform this comparison in two sets, the complete set of admissions, as well as the N2C2 subset for which also manually SDoH are available. For the N2C2 subset, we also assess the associations between the manually labeled SDoH and DNR/DNI status, acting as a reference standard for the text-mining based approaches. In the sections below we further specify the used text mining modeling details and the association study design.

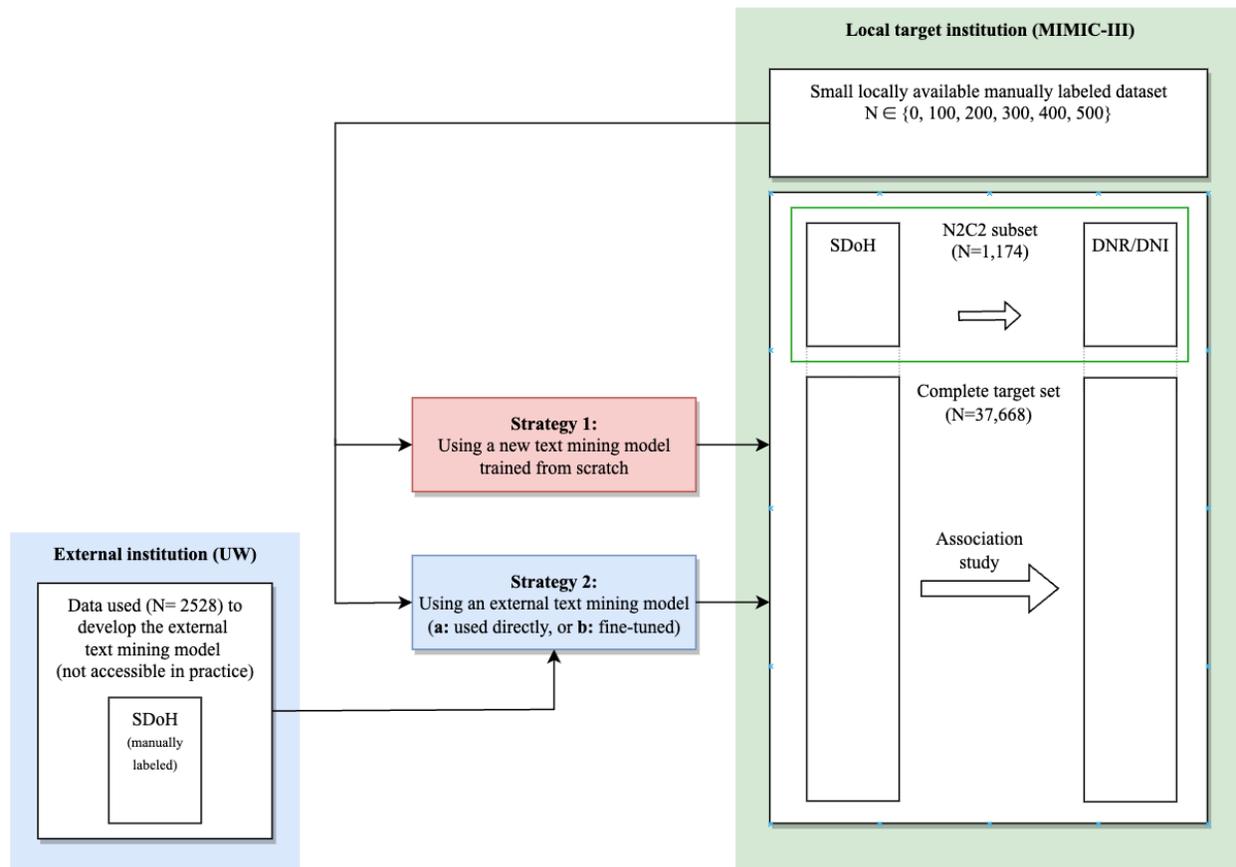

**Figure 2.** Schematic overview of the overall study design.

*The text mining models*

The text mining models extract information about employment status, living status, and substance use, including their timing (current, past, or future), specification (type of employment, and living status), as well as negations, following the labeling scheme used by the N2C2 shared task on SDoH extraction [12]. The model used was based on our submission[4] to the shared task, which performs the SDoH extraction via sequence tagging task in a BIO (Begin, In, Out) scheme, using Long-Short Term Memory (LSTM) networks with randomly initialized word embeddings[5]. Multiple text-mining models were trained for SDoH extraction in two different setups: (a) with transfer learning from a model trained on the complete set of manually labeled notes from University of Washington (UW) to the MIMIC-III manually labeled dataset, (b) training models from random initialization on the MIMIC-III manually labeled dataset. Five-fold cross-validation was used to estimate the overall performance on UW data before subsequent training on the entire UW set for transfer learning. The average overall F1-score across all the five folds was 0.68, indicating reasonable performance. A random subset of 500 training examples in the MIMIC-III dataset was further used to train models both from a random initialization (from scratch) and in a transfer learning setup. Subsequent variations of this subset were created with 400, 300, 200 and 100 training samples to analyze the impact of training data size. For transfer learning, the UW model was further trained for 250 epochs on the MIMIC-III subset using the same hyperparameter settings as our N2C2 shared task submission and a learning rate of 0.001[6]. The results for model performance on the manually labeled set used in the association study set is reported in **Figure 3**.

---

[4] https://github.com/tuur/sdoh_n2c2track2_ucsf_umcu/blob/main/N2C2%20Abstract.pdf
[5] We used this architecture among different variations because it performed well on the N2C2 shared tasks, was not dependent on embeddings from other data sources that may not always be available, and was the least resource intensive for future experiments.

*Extracting SDoH from the entire MIMIC-III corpus*

For the N2C2 subset, the social history sections were already separated by the shared task organizers, and were directly used in this study. The social history sections of discharge reports in the MIMIC-III corpus were first isolated from complete discharge reports with rule-based methods, specifically, an expansion of the available social history sectionizer in MedSpacy [48]. The text mining models were then applied to the social history information to extract the SDoH.

*The association study*

The five text-extracted studied SDoH are:

1. Living status (currently with family versus rest)
2. Employment status (currently employed versus rest)
3. Alcohol use (current or past use versus rest)
4. Drug use (current or past use versus rest)
5. Tobacco use (current or past use versus rest)

These studied SDoH were extracted from the N2C2 format outputted by the text mining models. For the living status and employment status, if there was no mention of employment or living status at all or if the identified timing was not current (but for example in the past) they were considered missing. The 'rest' category is assigned for living and employment status if a different current status was identified than the one studied. For alcohol use, drug use and tobacco use any mention of the event was considered, irrespective of timing (since we consider present or past use), unless an explicit negation was identified. The negated substance use statuses constitute the 'rest' categories.

The outcome is a commonly used composite [38,49]: having code status do not resuscitate or do not intubate (DNR/DNI) any time during the hospital stay. Code status was first collected from the structured data. Additional DNR/DNI codes were retrieved from the textual discharge summaries[7]. For each of the SDoH, the association with DNR/DNI was assessed using logistic regression, resulting in multivariable (log) odds ratios. Each association was adjusted for age, gender, ethnicity, and religion by adding these variables as covariates to the logistic regression model.

*Missing data*

Missing data were imputed using multiple imputation by chained equations (MICE) with predictive mean matching [51]. Ten imputation sets were used, analyses were performed in each imputation set separately, and resulting associations were pooled across imputation sets using Rubin's rules [52]. The variables used in the imputation model are: age, gender, ethnicity, religion, marital status, admission location, insurance, admission type, the studied outcome, and the SDoH.

---

[6] We selected 0.001 as it is the default learning rate for Adam optimizer.
[7] If there was mention of DNR or DNI variants in the discharge summary - and no full code was mentioned.

*Implementation*

All analyses were performed in Python 3. The code used for the entire study, as well as exact links to the public MIMIC-III database are available on https://github.com/tuur/sdoh_n2c2track2_ucsf_umcu.

# Results

*Patient characteristics*

Excluding the random 500 patients used for training and fine-tuning the text mining models, 37,668 patient admissions met the inclusion criteria, of which 1,174 had manually labeled SDoH available via the N2C2 shared task. Although the N2C2 subset is not a random subset, as it was manually labeled via active learning [12], the observed characteristics used in our analyses are similar, observable from Table 1. Overall the prevalence of DNR/DNI status was 8%. It can be observed from the N2C2 subset that for the manually labeled text-based patient characteristics, there are quite some missings in the SDoH information, ranging from 24% (alcohol status) to 54% (employment status) missingness, due to absence of reporting SDoH information in the clinical notes.

**Table 1.** Patient characteristics for the data used for the association studies. Characteristics labeled in free text are indicated with a asterisk (*). These values are only available for the N2C2 subset.

|  |  | N2C2 MIMIC-III subset (n=1,174) | MIMIC-III (n=37,668) |
|---|---|---|---|
| Age, median (IQR) |  | 61 [50, 73] | 64 [52, 76] |
| Gender, n (%) | Female | 463 (39.4) | 16,179 (43.0) |
|  | Male | 711 (60.6) | 21,489 (57.0) |
| Admission type, n (%) | Elective | 136 (11.6) | 5,476 (14.5) |
|  | Emergency | 1,017 (86.6) | 31,430 (83.4) |
|  | Urgent | 21 (1.8) | 762 (2.0) |
| DNR/DNI, n (%) | No | 1,075 (91.6) | 34,640 (92.0) |
|  | Yes | 99 (8.4) | 3,028 (8.0) |
| Employment*, n (%) | Employed | 248 (21.1) | - |
|  | Other | 293 (25.0) | - |
|  | Missing | 633 (53.9) | - |
| Tobacco*, n (%) | Current/past | 604 (51.4) | - |
|  | Other | 245 (20.9) | - |
|  | Missing | 325 (27.7) | - |
| Alcohol*, n (%) | Current/past | 615 (52.4) | - |
|  | Other | 275 (23.4) | - |
|  | Missing | 284 (24.2) | - |
| Drug status*, n (%) | Current/past | 240 (20.4) | - |
|  | Other | 362 (30.8) | - |
| | Missing | 572 (48.7) | - |
| Living status*, n (%) | With family | 373 (31.8) | - |
| | Other | 192 (16.4) | - |
| | Missing | 609 (51.9) | - |
| Ethnicity, n (%) | Asian | 12 (1.0) | 352 (0.9) |
| | Black | 124 (10.6) | 3,648 (9.7) |
| | Hispanic | 47 (4.0) | 1,340 (3.6) |
| | Not specified | 100 (8.5) | 3,322 (8.8) |
| | Other | 50 (4.3) | 1,994 (5.3) |
| | White | 841 (71.6) | 27,012 (71.7) |
| Religion, n (%) | Catholic | 450 (38.3) | 14,041 (37.3) |
| | Jewish | 74 (6.3) | 3,048 (8.1) |
| | Not specified | 400 (34.1) | 12,143 (32.2) |
| | Other | 250 (21.3) | 8,436 (22.4) |

*Text mining model performance*

The text mining model F1-score ranged between 0.57 and 0.65 on the N2C2 MIMIC-III subset used for the association study (**Figure 3**; detailed results for each SDoH factor provided in **Supplementary Materials, Tables A1-A11**). The model that was first trained on UW data and then fine-tuned further on MIMIC-III examples performed better than models directly trained on MIMIC-III examples, likely due to the larger sample size used in transfer learning. The precision scores were numerically higher, but did not change much across different sample sizes. The trend of F1-score was driven by an increase in the recall values across the increasing number of samples.

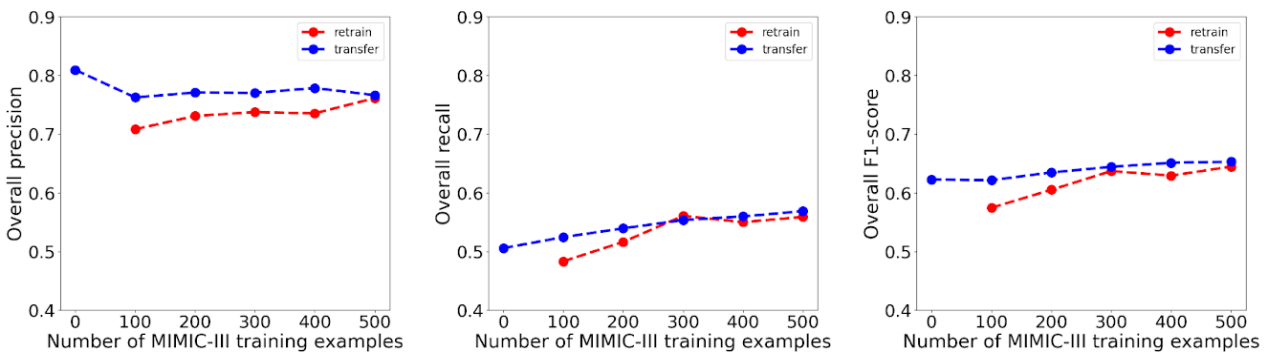

**Figure 3.** Text mining performance, in terms of overall precision (true positive rate), overall recall (sensitivity) and overall $F_1$-score (harmonic mean of precision and recall) on the N2C2 MIMIC-III subset used for the association study when: a) fine-tuning on an increasing amount of within-institution (MIMIC-III) data by transferring a model trained on external (UW) data (in blue), and b) training a new model only on within-institution (MIMIC-III) data (in red).

*Association study results*

In the N2C2 subset results (**Figure 4**), it can be observed that the confidence intervals are quite wide due to the limited amount of data. For living status, and alcohol use the text-mining-based associations are quite close to those found using manually labeled data. However, for employment status, drugs use, and tobacco use the text-mining-based associations were quite different from the association estimates obtained using manually labeled data. This difference was especially large for employment status, which could be driven by a relatively lower F1-measure of the text mining models for employment trigger extraction (nearly 10% lower compared to the other SDoH, **Supplementary materials Tables A1-A11**). A peak is seen in the odds ratio for employment status when using 300 samples, which could be related to model instability coming from training with a relatively small sample size. Comparing the different text-mining setups, no meaningful differences in the associations were found between any of the methods: the external UW model, the fine-tuned models, nor the models trained from scratch. The same analyses based only on the complete-case data (**Supplementary materials Figure A1-CCA**) show similar results.

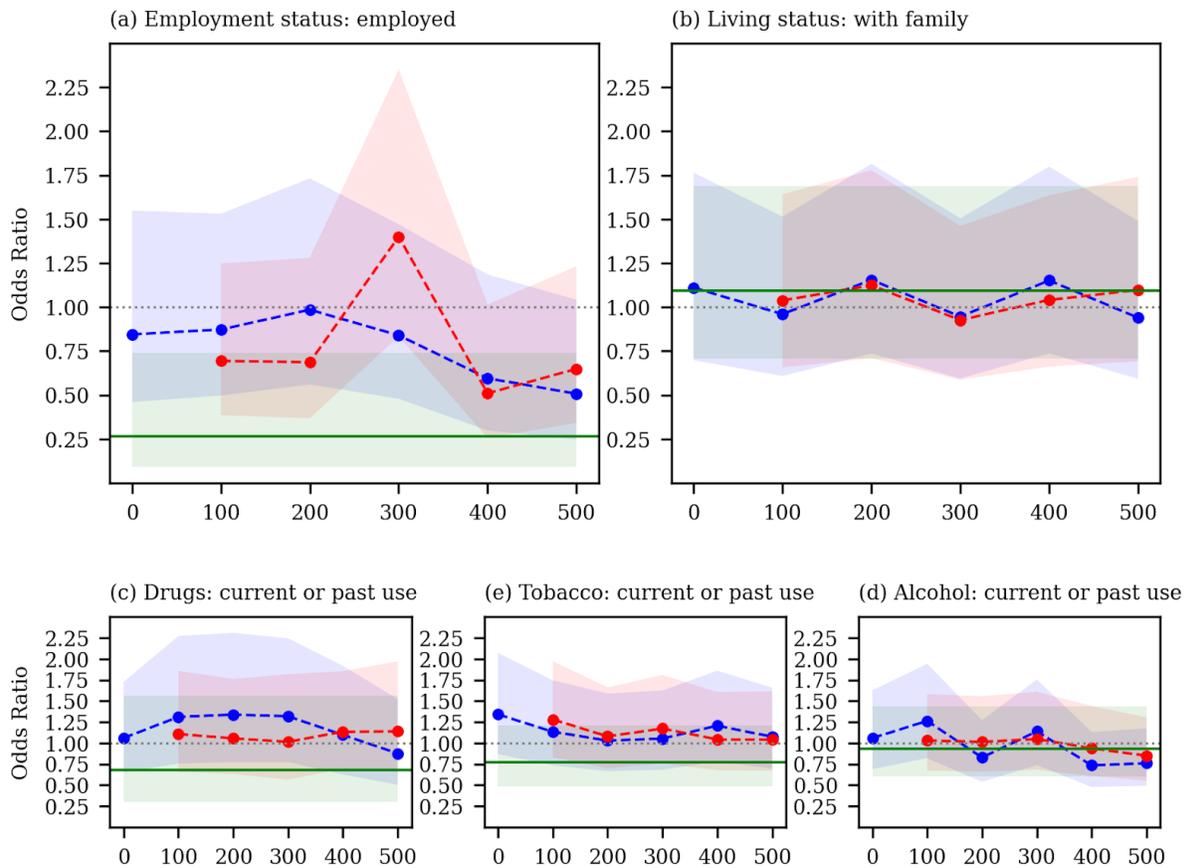

**Figure 4**. Associations (with 95% confidence intervals) on the **1174 pt. manually labeled subset** of MIMIC-III for varying sizes of manually labeled data (x-axis) used for text mining fine tuning starting from the UW model (in blue) or training a new model only on within-institution data (in red). The association found using manually labeled data is indicated by the horizontal line (in green).

When inspecting the results on the larger section of MIMIC-III (**Figure 5**), it can be observed that the confidence intervals are much more narrow, due to more data being used. Again comparing the different text-mining setups, there were no meaningful differences in the found associations between the external UW model, the fine-tuned models, nor the new models trained from scratch. While in the N2C2 subset a clear negative text-mining-based association was found only for being currently employed, in the larger MIMIC-III set, a clear negative association was additionally found between living with family and DNR/DNI code, and a clear positive association was found between past or current drug use and DNR/DNI code. The reliability of these text-mining-based associations could not be confirmed on the larger MIMIC-III set, as no labeled data is available. Performing these analyses only on the complete-case data (**Supplementary materials, Figure A2-CCA**) show similar results, except that when using the UW model without any fine tuning, no negative association was found between living status and having a DNR/DNI code.

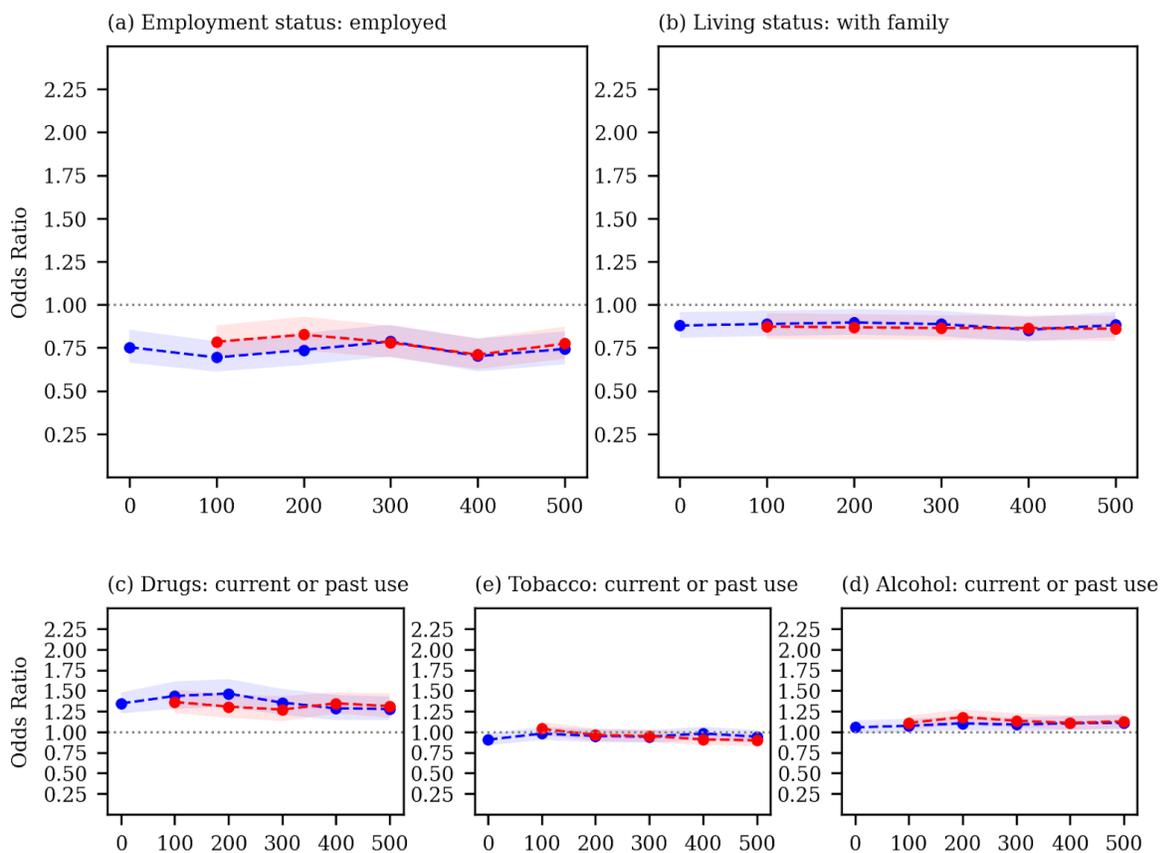

**Figure 5**. Associations (with 95% confidence intervals) on the larger **37,668 pt.** MIMIC-III data for varying sizes of labeled data (x-axis) used for text mining fine tuning starting from the UW model (in blue) or training a new model only on within-institution data (in red).

**Discussion**

In our study, we analyzed the associations between being employed, living with family, and history of substance use with the patients' DNR/DNI code status at any time during an admission. We further compared the differences between the associations when using different text mining models to extract the SDoH factors. Important differences were found between different text mining models and the manually labeled information for three out of four studied factors. This suggests that the currently used text mining models may not be reliable enough to conduct a multivariable association study. Furthermore, using an increasing sample size for training and fine-tuning text mining models did not show meaningfully different association estimates. This could be so because the differences in F1 scores of the models were not large enough for driving differences in the identified associations. While with the developed text mining models we were able to extract SDoH factors from a large dataset, leading to smaller confidence intervals than for the manually labeled subset provided by the N2C2 shared task (**Figure 4** vs. **Figure 5**), we could not confirm the reliability of these estimates due to the absence of manually added SDoH labels. Development of better text mining models may improve the reliability of the associations, which should ideally be confirmed on a larger manually labeled dataset. This should be addressed in future research.

To further appreciate our study, several issues need to be addressed, related to the use of electronic medical record data. While the free text discharge summaries provided information about SDoH that were not structurally recorded and would have otherwise been ignored, the SDoH were not recorded in the textual notes for *all* patients. This may have impacted the identified associations, as the missingness may be related to unmeasured factors ('missing not at random' [53]) that were not included in the used imputation models, which handles only missingness related to the measured variables [54]. Finally, code status was not structurally recorded for all patients, which led us to extract DNR/DNI codes also from textual notes. Nevertheless, we may still have missed patients that had a DNR/DNI code some time during their stay. Finally, we did not compare the characteristics of the patients at the external institution with the local target institution, as we did not have access to this data. While this is not ideal, in our view, it can be expected in practice due to data privacy concerns.

Beside these considerations about data quality, certain methodological issues should be addressed. In the current study, we adapted the externally trained model on manually labeled examples from the local institution by using the default learning rate settings for the optimizer. While the model's overall F1 score showed a clear improvement when adding more local data, further improvement may have been possible by fine-tuning the hyperparameters towards the local dataset. Furthermore, while the N2C2 dataset manually labeled with SDoH information provided the necessary cross-institution data for training and validation of the text mining models, the active learning scheme that was used to sample the reports to label may have influenced the results. Although we obtain similar summary statistics for the N2C2 subset compared to the complete MIMIC-III dataset (**Table 1**), the effectiveness of the retraining and finetuning process, the representativeness of the N2C2 subset, as well as the associations found from the manually labeled data may still be impacted by the sampling strategy used for labeling.

**Conclusions**

For three out of the five studied social factors, the text-mining based associations differed meaningfully from the associations based on manually labeled data. This indicates that the text mining models were not sufficiently accurate to reliably be used in multivariable association studies. To better study the relation between text mining performance and reliability of consequent medical studies, future studies using more manually labeled data are needed. No meaningful differences were found between fine tuning an external model on within-institution data or retraining a new model from scratch in any of the experiments. Future studies should explore better text mining models and confirm the reliability of these models for downstream association studies on a larger manually labeled dataset. It should focus on identifying the conditions in which a text mining model could be used sufficiently reliably to assess multivariable associations.

## Acknowledgements

We thank the organizers of the 2022 N2C2 Shared Task on Challenges in NLP for Clinical Data for their provision of the data, and K. Lybarger for providing the alignment between the shared task and the MIMIC-III data. We would additionally like to thank the Wynton team at University of California, San

Francisco for supporting the high-performance computational infrastructure that was used to train several models used in this study.

**Competing interests**

AJB is a co-founder and consultant to Personalis and NuMedii; consultant to Mango Tree Corporation, and in the recent past, Samsung, 10x Genomics, Helix, Pathway Genomics, and Verinata (Illumina); has served on paid advisory panels or boards for Geisinger Health, Regenstrief Institute, Gerson Lehman Group, AlphaSights, Covance, Novartis, Genentech, and Merck, and Roche; is a shareholder in Personalis and NuMedii; is a minor shareholder in Apple, Meta (Facebook), Alphabet (Google), Microsoft, Amazon, Snap, 10x Genomics, Illumina, Regeneron, Sanofi, Pfizer, Royalty Pharma, Moderna, Sutro, Doximity, BioNtech, Invitae, Pacific Biosciences, Editas Medicine, Nuna Health, Assay Depot, and Vet24seven, and several other non-health related companies and mutual funds; and has received honoraria and travel reimbursement for invited talks from Johnson and Johnson, Roche, Genentech, Pfizer, Merck, Lilly, Takeda, Varian, Mars, Siemens, Optum, Abbott, Celgene, AstraZeneca, AbbVie, Westat, and many academic institutions, medical or disease specific foundations and associations, and health systems. AJB receives royalty payments through Stanford University, for several patents and other disclosures licensed to NuMedii and Personalis. AJB's research has been funded by NIH, Peraton (as the prime on an NIH contract), Genentech, Johnson and Johnson, FDA, Robert Wood Johnson Foundation, Leon Lowenstein Foundation, Intervalien Foundation, Priscilla Chan and Mark Zuckerberg, the Barbara and Gerson Bakar Foundation, and in the recent past, the March of Dimes, Juvenile Diabetes Research Foundation, California Governor's Office of Planning and Research, California Institute for Regenerative Medicine, L'Oreal, and Progenity.  The authors have declared that no competing interests exist.

**Author contributions**

MS and AL were responsible for the conceptualization and design of the study, performing the analyses, and drafting the original manuscript. AJB, MS, and ES critically analyzed, reviewed, contributed to, and approved the final manuscript.

# Supplementary file A

## Complete case analysis (CCA) results

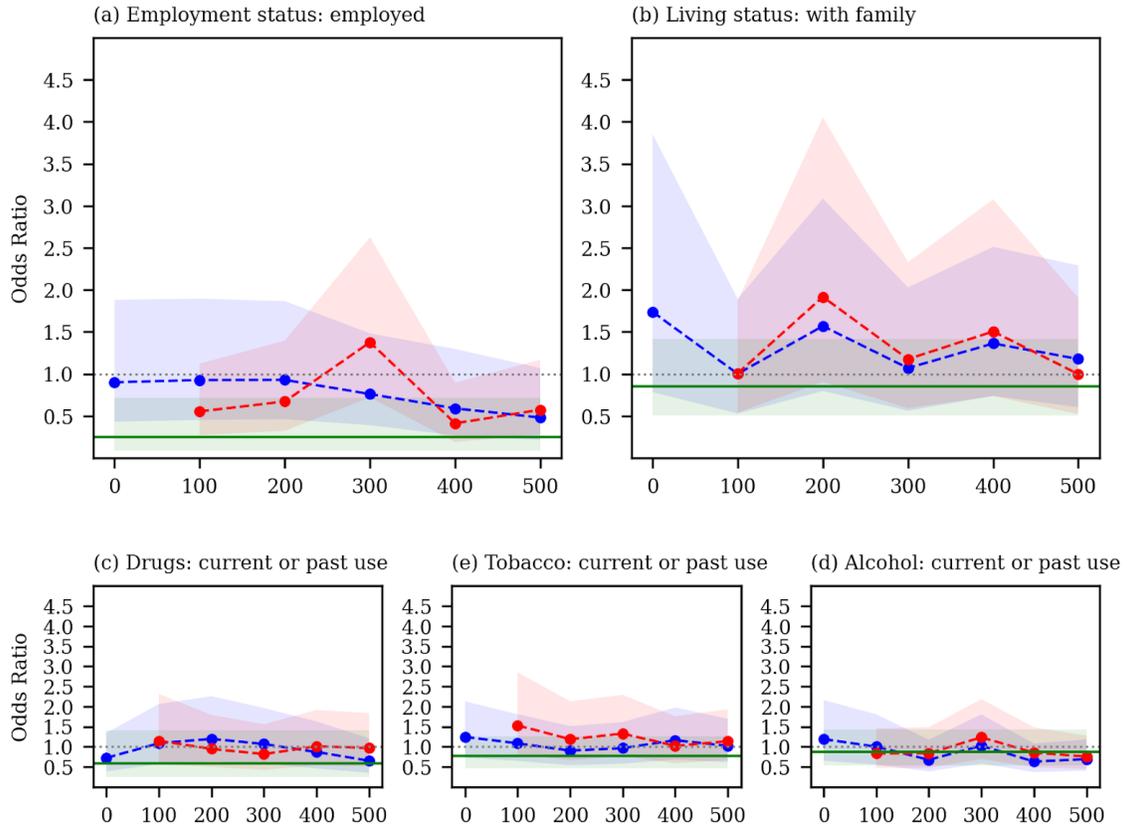

**Figure A1-CCA**. Associations (with their 95% confidence intervals) on the 1174 pt. manually labeled subset of MIMIC for varying sizes of manually labeled data (x-axis) used for model text mining model fine tuning starting from the UW model (in blue) or training a new model only on within-insitution data (in red). The association found using manually labeled data is indicated by the horizontal line (in green).

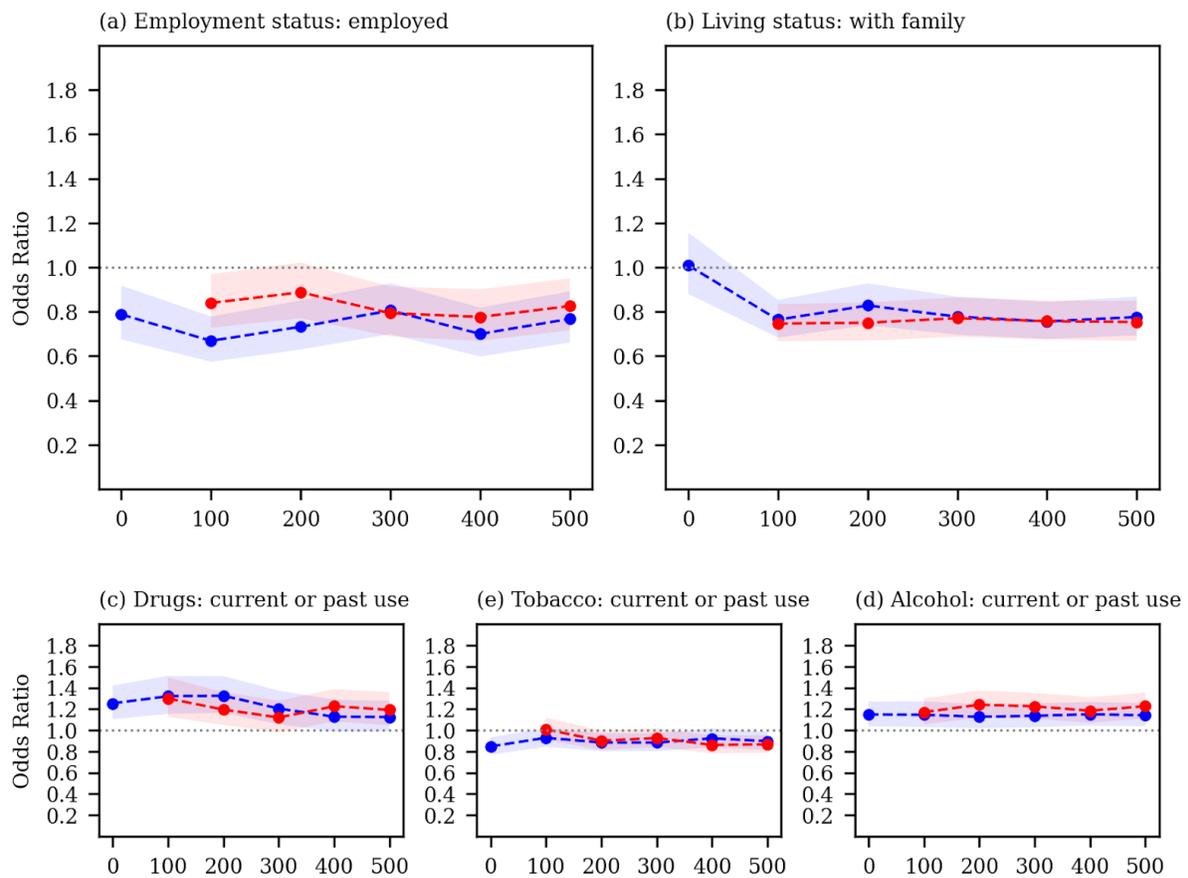

**Figure A2-CCA**. Associations (with 95% confidence intervals) on the **37.668 pt. unlabeled subset** of MIMIC for varying sizes of manually labeled data (x-axis) used for text mining fine tuning starting from the UW model (in blue) or training a new model only on within-institution data (in red).

# Detailed SDoH extraction results for all the models used for text mining

Table A1: Precision, recall and F1 score obtained when directly applying the SDoH information extraction model trained only on **UW data.** NT is the count of true (gold) labels, NP is the count of predicted labels, TP is the counted true positives, P is precision, R is recall and F1 is the F-1 measure, which is the harmonic mean of precision and recall.

| SDoH event | Argument | Subtype | NT | NP | TP | P | R | F1 |
|---|---|---|---|---|---|---|---|---|
| OVERALL | OVERALL | OVERALL | 13110 | 8190 | 6629 | 0.81 | 0.51 | 0.62 |
| Alcohol | Amount | N/A | 236 | 28 | 18 | 0.64 | 0.08 | 0.14 |
| Alcohol | Duration | N/A | 52 | 0 | 0 | 0.00 | 0.00 | 0.00 |
| Alcohol | Frequency | N/A | 233 | 66 | 32 | 0.48 | 0.14 | 0.21 |
| Alcohol | History | N/A | 116 | 1 | 0 | 0.00 | 0.00 | 0.00 |
| Alcohol | StatusTime | current | 460 | 239 | 166 | 0.69 | 0.36 | 0.47 |
| Alcohol | StatusTime | none | 391 | 278 | 232 | 0.83 | 0.59 | 0.69 |
| Alcohol | StatusTime | past | 229 | 15 | 10 | 0.67 | 0.04 | 0.08 |
| Alcohol | Trigger | N/A | 1080 | 1030 | 873 | 0.85 | 0.81 | 0.83 |
| Alcohol | Type | N/A | 140 | 59 | 23 | 0.39 | 0.16 | 0.23 |
| Drug | Amount | N/A | 5 | 0 | 0 | 0.00 | 0.00 | 0.00 |
| Drug | Duration | N/A | 13 | 0 | 0 | 0.00 | 0.00 | 0.00 |
| Drug | Frequency | N/A | 21 | 0 | 0 | 0.00 | 0.00 | 0.00 |
| Drug | History | N/A | 71 | 0 | 0 | 0.00 | 0.00 | 0.00 |
| Drug | Method | N/A | 146 | 130 | 66 | 0.51 | 0.45 | 0.48 |
| Drug | StatusTime | current | 128 | 30 | 15 | 0.50 | 0.12 | 0.19 |
| Drug | StatusTime | none | 491 | 233 | 200 | 0.86 | 0.41 | 0.55 |
| Drug | StatusTime | past | 150 | 22 | 11 | 0.50 | 0.07 | 0.13 |
| Drug | Trigger | N/A | 769 | 765 | 648 | 0.85 | 0.84 | 0.84 |
| Drug | Type | N/A | 494 | 322 | 231 | 0.72 | 0.47 | 0.57 |
| Employment | Duration | N/A | 30 | 0 | 0 | 0.00 | 0.00 | 0.00 |
| Employment | History | N/A | 35 | 0 | 0 | 0.00 | 0.00 | 0.00 |
| Employment | StatusEmploy | employed | 223 | 135 | 108 | 0.80 | 0.48 | 0.60 |
| Employment | StatusEmploy | homemaker | 9 | 0 | 0 | 0.00 | 0.00 | 0.00 |
| Employment | StatusEmploy | on_disability | 102 | 54 | 40 | 0.74 | 0.39 | 0.51 |
| Employment | StatusEmploy | retired | 141 | 107 | 94 | 0.88 | 0.67 | 0.76 |
| Employment | StatusEmploy | student | 13 | 4 | 3 | 0.75 | 0.23 | 0.35 |
| Employment | StatusEmploy | unemployed | 275 | 148 | 104 | 0.70 | 0.38 | 0.49 |
| Employment | Trigger | N/A | 763 | 605 | 485 | 0.80 | 0.64 | 0.71 |
| Employment | Type | N/A | 559 | 81 | 30 | 0.37 | 0.05 | 0.09 |
| LivingStatus | Duration | N/A | 23 | 0 | 0 | 0.00 | 0.00 | 0.00 |
| LivingStatus | History | N/A | 10 | 0 | 0 | 0.00 | 0.00 | 0.00 |

| SDoH event | Argument | Subtype | NT | NP | TP | P | R | F1 |
|---|---|---|---|---|---|---|---|---|
| LivingStatus | StatusTime | current | 761 | 688 | 624 | 0.91 | 0.82 | 0.86 |
| LivingStatus | StatusTime | future | 1 | 0 | 0 | 0.00 | 0.00 | 0.00 |
| LivingStatus | StatusTime | none | 1 | 0 | 0 | 0.00 | 0.00 | 0.00 |
| LivingStatus | StatusTime | past | 49 | 1 | 1 | 1.00 | 0.02 | 0.04 |
| LivingStatus | Trigger | N/A | 812 | 796 | 698 | 0.88 | 0.86 | 0.87 |
| LivingStatus | TypeLiving | alone | 169 | 96 | 86 | 0.90 | 0.51 | 0.65 |
| LivingStatus | TypeLiving | homeless | 33 | 32 | 24 | 0.75 | 0.73 | 0.74 |
| LivingStatus | TypeLiving | with_family | 447 | 349 | 309 | 0.89 | 0.69 | 0.78 |
| LivingStatus | TypeLiving | with_others | 163 | 64 | 33 | 0.52 | 0.20 | 0.29 |
| Tobacco | Amount | N/A | 445 | 176 | 111 | 0.63 | 0.25 | 0.36 |
| Tobacco | Duration | N/A | 215 | 30 | 20 | 0.67 | 0.09 | 0.16 |
| Tobacco | Frequency | N/A | 141 | 63 | 42 | 0.67 | 0.30 | 0.41 |
| Tobacco | History | N/A | 266 | 43 | 22 | 0.51 | 0.08 | 0.14 |
| Tobacco | Method | N/A | 3 | 0 | 0 | 0.00 | 0.00 | 0.00 |
| Tobacco | StatusTime | current | 312 | 140 | 98 | 0.70 | 0.31 | 0.43 |
| Tobacco | StatusTime | none | 355 | 235 | 213 | 0.91 | 0.60 | 0.72 |
| Tobacco | StatusTime | past | 383 | 210 | 167 | 0.80 | 0.44 | 0.56 |
| Tobacco | Trigger | N/A | 1050 | 882 | 777 | 0.88 | 0.74 | 0.80 |
| Tobacco | Type | N/A | 96 | 33 | 15 | 0.45 | 0.16 | 0.23 |

**Table A2:** Precision, recall and F1 score obtained when adapting the SDoH information extraction model from **UW data with 100 samples from the MIMIC-III train set.** NT is the count of true (gold) labels, NP is the count of predicted labels, TP is the counted true positives, P is precision, R is recall and F1 is the F-1 measure, which is the harmonic mean of precision and recall.

| SDoH event | Argument | Subtype | NT | NP | TP | P | R | F1 |
|---|---|---|---|---|---|---|---|---|
| OVERALL | OVERALL | OVERALL | 13110 | 9016 | 6874 | 0.76 | 0.52 | 0.62 |
| Alcohol | Amount | N/A | 236 | 55 | 24 | 0.44 | 0.10 | 0.16 |
| Alcohol | Duration | N/A | 52 | 0 | 0 | 0.00 | 0.00 | 0.00 |
| Alcohol | Frequency | N/A | 233 | 145 | 66 | 0.46 | 0.28 | 0.35 |
| Alcohol | History | N/A | 116 | 1 | 0 | 0.00 | 0.00 | 0.00 |
| Alcohol | StatusTime | current | 460 | 240 | 150 | 0.63 | 0.33 | 0.43 |
| Alcohol | StatusTime | none | 391 | 337 | 261 | 0.77 | 0.67 | 0.72 |
| Alcohol | StatusTime | past | 229 | 48 | 25 | 0.52 | 0.11 | 0.18 |
| Alcohol | Trigger | N/A | 1080 | 1063 | 875 | 0.82 | 0.81 | 0.82 |
| Alcohol | Type | N/A | 140 | 113 | 44 | 0.39 | 0.31 | 0.35 |
| Drug | Amount | N/A | 5 | 0 | 0 | 0.00 | 0.00 | 0.00 |
| Drug | Duration | N/A | 13 | 0 | 0 | 0.00 | 0.00 | 0.00 |
| Drug | Frequency | N/A | 21 | 0 | 0 | 0.00 | 0.00 | 0.00 |
| Drug | History | N/A | 71 | 0 | 0 | 0.00 | 0.00 | 0.00 |
| Drug | Method | N/A | 146 | 95 | 49 | 0.52 | 0.34 | 0.41 |
| Drug | StatusTime | current | 128 | 14 | 9 | 0.64 | 0.07 | 0.13 |
| Drug | StatusTime | none | 491 | 376 | 291 | 0.77 | 0.59 | 0.67 |
| Drug | StatusTime | past | 150 | 16 | 7 | 0.44 | 0.05 | 0.08 |
| Drug | Trigger | N/A | 769 | 703 | 584 | 0.83 | 0.76 | 0.79 |
| Drug | Type | N/A | 494 | 349 | 220 | 0.63 | 0.45 | 0.52 |
| Employment | Duration | N/A | 30 | 0 | 0 | 0.00 | 0.00 | 0.00 |
| Employment | History | N/A | 35 | 0 | 0 | 0.00 | 0.00 | 0.00 |
| Employment | StatusEmploy | employed | 223 | 131 | 116 | 0.89 | 0.52 | 0.66 |
| Employment | StatusEmploy | homemaker | 9 | 0 | 0 | 0.00 | 0.00 | 0.00 |
| Employment | StatusEmploy | on_disability | 102 | 47 | 36 | 0.77 | 0.35 | 0.48 |
| Employment | StatusEmploy | retired | 141 | 108 | 92 | 0.85 | 0.65 | 0.74 |
| Employment | StatusEmploy | student | 13 | 3 | 3 | 1.00 | 0.23 | 0.38 |
| Employment | StatusEmploy | unemployed | 275 | 181 | 104 | 0.57 | 0.38 | 0.46 |
| Employment | Trigger | N/A | 763 | 669 | 504 | 0.75 | 0.66 | 0.70 |
| Employment | Type | N/A | 559 | 121 | 37 | 0.31 | 0.07 | 0.11 |
| LivingStatus | Duration | N/A | 23 | 0 | 0 | 0.00 | 0.00 | 0.00 |
| LivingStatus | History | N/A | 10 | 0 | 0 | 0.00 | 0.00 | 0.00 |
| LivingStatus | StatusTime | current | 761 | 760 | 662 | 0.87 | 0.87 | 0.87 |
| LivingStatus | StatusTime | future | 1 | 0 | 0 | 0.00 | 0.00 | 0.00 |
| LivingStatus | StatusTime | none | 1 | 0 | 0 | 0.00 | 0.00 | 0.00 |
| LivingStatus | StatusTime | past | 49 | 2 | 2 | 1.00 | 0.04 | 0.08 |
| LivingStatus | Trigger | N/A | 812 | 783 | 695 | 0.89 | 0.86 | 0.87 |
| LivingStatus | TypeLiving | alone | 169 | 142 | 129 | 0.91 | 0.76 | 0.83 |
| LivingStatus | TypeLiving | homeless | 33 | 11 | 9 | 0.82 | 0.27 | 0.41 |
| LivingStatus | TypeLiving | with_family | 447 | 385 | 343 | 0.89 | 0.77 | 0.82 |
| LivingStatus | TypeLiving | with_others | 163 | 86 | 42 | 0.49 | 0.26 | 0.34 |
| Tobacco | Amount | N/A | 445 | 213 | 121 | 0.57 | 0.27 | 0.37 |
| Tobacco | Duration | N/A | 215 | 55 | 25 | 0.45 | 0.12 | 0.19 |
| Tobacco | Frequency | N/A | 141 | 80 | 44 | 0.55 | 0.31 | 0.40 |
| Tobacco | History | N/A | 266 | 116 | 51 | 0.44 | 0.19 | 0.27 |
| Tobacco | Method | N/A | 3 | 0 | 0 | 0.00 | 0.00 | 0.00 |
| Tobacco | StatusTime | current | 312 | 143 | 91 | 0.64 | 0.29 | 0.40 |
| Tobacco | StatusTime | none | 355 | 200 | 171 | 0.86 | 0.48 | 0.62 |
| Tobacco | StatusTime | past | 383 | 210 | 156 | 0.74 | 0.41 | 0.53 |
| Tobacco | Trigger | N/A | 1050 | 953 | 809 | 0.85 | 0.77 | 0.81 |
| Tobacco | Type | N/A | 96 | 62 | 27 | 0.44 | 0.28 | 0.34 |

**Table A3:** Precision, recall and F1 score obtained when adapting the SDoH information extraction model from **UW data with 200 samples from the MIMIC-III train set.** NT is the count of true (gold) labels, NP is the count of predicted labels, TP is the counted true positives, P is precision, R is recall and F1 is the F-1 measure, which is the harmonic mean of precision and recall.

| SDoH event | Argument | Subtype | NT | NP | TP | P | R | F1 |
|---|---|---|---|---|---|---|---|---|
| OVERALL | OVERALL | OVERALL | 13110 | 9172 | 7070 | 0.77 | 0.54 | 0.63 |
| Alcohol | Amount | N/A | 236 | 39 | 17 | 0.44 | 0.07 | 0.12 |
| Alcohol | Duration | N/A | 52 | 0 | 0 | 0.00 | 0.00 | 0.00 |
| Alcohol | Frequency | N/A | 233 | 85 | 34 | 0.40 | 0.15 | 0.21 |
| Alcohol | History | N/A | 116 | 0 | 0 | 0.00 | 0.00 | 0.00 |
| Alcohol | StatusTime | current | 460 | 248 | 160 | 0.65 | 0.35 | 0.45 |
| Alcohol | StatusTime | none | 391 | 320 | 247 | 0.77 | 0.63 | 0.69 |
| Alcohol | StatusTime | past | 229 | 34 | 19 | 0.56 | 0.08 | 0.14 |
| Alcohol | Trigger | N/A | 1080 | 1043 | 866 | 0.83 | 0.80 | 0.82 |
| Alcohol | Type | N/A | 140 | 118 | 48 | 0.41 | 0.34 | 0.37 |
| Drug | Amount | N/A | 5 | 0 | 0 | 0.00 | 0.00 | 0.00 |
| Drug | Duration | N/A | 13 | 0 | 0 | 0.00 | 0.00 | 0.00 |
| Drug | Frequency | N/A | 21 | 0 | 0 | 0.00 | 0.00 | 0.00 |
| Drug | History | N/A | 71 | 0 | 0 | 0.00 | 0.00 | 0.00 |
| Drug | Method | N/A | 146 | 81 | 45 | 0.56 | 0.31 | 0.40 |
| Drug | StatusTime | current | 128 | 14 | 6 | 0.43 | 0.05 | 0.08 |
| Drug | StatusTime | none | 491 | 409 | 319 | 0.78 | 0.65 | 0.71 |
| Drug | StatusTime | past | 150 | 20 | 8 | 0.40 | 0.05 | 0.09 |
| Drug | Trigger | N/A | 769 | 769 | 647 | 0.84 | 0.84 | 0.84 |
| Drug | Type | N/A | 494 | 409 | 264 | 0.65 | 0.53 | 0.58 |
| Employment | Duration | N/A | 30 | 0 | 0 | 0.00 | 0.00 | 0.00 |
| Employment | History | N/A | 35 | 0 | 0 | 0.00 | 0.00 | 0.00 |
| Employment | StatusEmploy | employed | 223 | 152 | 123 | 0.81 | 0.55 | 0.66 |
| Employment | StatusEmploy | homemaker | 9 | 0 | 0 | 0.00 | 0.00 | 0.00 |
| Employment | StatusEmploy | on_disability | 102 | 43 | 35 | 0.81 | 0.34 | 0.48 |
| Employment | StatusEmploy | retired | 141 | 120 | 98 | 0.82 | 0.70 | 0.75 |
| Employment | StatusEmploy | student | 13 | 3 | 3 | 1.00 | 0.23 | 0.38 |
| Employment | StatusEmploy | unemployed | 275 | 188 | 118 | 0.63 | 0.43 | 0.51 |
| Employment | Trigger | N/A | 763 | 672 | 520 | 0.77 | 0.68 | 0.72 |
| Employment | Type | N/A | 559 | 152 | 57 | 0.38 | 0.10 | 0.16 |
| LivingStatus | Duration | N/A | 23 | 0 | 0 | 0.00 | 0.00 | 0.00 |
| LivingStatus | History | N/A | 10 | 0 | 0 | 0.00 | 0.00 | 0.00 |
| LivingStatus | StatusTime | current | 761 | 759 | 657 | 0.87 | 0.86 | 0.86 |

| SDoH event | Argument | Subtype | NT | NP | TP | P | R | F1 |
|---|---|---|---|---|---|---|---|---|
| LivingStatus | StatusTime | future | 1 | 0 | 0 | 0.00 | 0.00 | 0.00 |
| LivingStatus | StatusTime | none | 1 | 0 | 0 | 0.00 | 0.00 | 0.00 |
| LivingStatus | StatusTime | past | 49 | 0 | 0 | 0.00 | 0.00 | 0.00 |
| LivingStatus | Trigger | N/A | 812 | 795 | 703 | 0.88 | 0.87 | 0.87 |
| LivingStatus | TypeLiving | alone | 169 | 150 | 133 | 0.89 | 0.79 | 0.83 |
| LivingStatus | TypeLiving | homeless | 33 | 14 | 13 | 0.93 | 0.39 | 0.55 |
| LivingStatus | TypeLiving | with_family | 447 | 364 | 330 | 0.91 | 0.74 | 0.81 |
| LivingStatus | TypeLiving | with_others | 163 | 88 | 45 | 0.51 | 0.28 | 0.36 |
| Tobacco | Amount | N/A | 445 | 235 | 118 | 0.50 | 0.27 | 0.35 |
| Tobacco | Duration | N/A | 215 | 17 | 11 | 0.65 | 0.05 | 0.09 |
| Tobacco | Frequency | N/A | 141 | 96 | 55 | 0.57 | 0.39 | 0.46 |
| Tobacco | History | N/A | 266 | 136 | 66 | 0.49 | 0.25 | 0.33 |
| Tobacco | Method | N/A | 3 | 0 | 0 | 0.00 | 0.00 | 0.00 |
| Tobacco | StatusTime | current | 312 | 148 | 98 | 0.66 | 0.31 | 0.43 |
| Tobacco | StatusTime | none | 355 | 222 | 195 | 0.88 | 0.55 | 0.68 |
| Tobacco | StatusTime | past | 383 | 250 | 187 | 0.75 | 0.49 | 0.59 |
| Tobacco | Trigger | N/A | 1050 | 955 | 814 | 0.85 | 0.78 | 0.81 |
| Tobacco | Type | N/A | 96 | 24 | 11 | 0.46 | 0.11 | 0.18 |

**Table A4:** Precision, recall and F1 score obtained when adapting the SDoH information extraction model from **UW data with 300 samples from the MIMIC-III train set.** NT is the count of true (gold) labels, NP is the count of predicted labels, TP is the counted true positives, P is precision, R is recall and F1 is the F-1 measure, which is the harmonic mean of precision and recall.

| SDoH event | Argument | Subtype | NT | NP | TP | P | R | F1 |
|---|---|---|---|---|---|---|---|---|
| OVERALL | OVERALL | OVERALL | 13110 | 9420 | 7256 | 0.77 | 0.55 | 0.64 |
| Alcohol | Amount | N/A | 236 | 33 | 16 | 0.48 | 0.07 | 0.12 |
| Alcohol | Duration | N/A | 52 | 0 | 0 | 0.00 | 0.00 | 0.00 |
| Alcohol | Frequency | N/A | 233 | 127 | 52 | 0.41 | 0.22 | 0.29 |
| Alcohol | History | N/A | 116 | 0 | 0 | 0.00 | 0.00 | 0.00 |
| Alcohol | StatusTime | current | 460 | 276 | 167 | 0.61 | 0.36 | 0.45 |
| Alcohol | StatusTime | none | 391 | 320 | 250 | 0.78 | 0.64 | 0.70 |
| Alcohol | StatusTime | past | 229 | 65 | 34 | 0.52 | 0.15 | 0.23 |
| Alcohol | Trigger | N/A | 1080 | 1101 | 896 | 0.81 | 0.83 | 0.82 |
| Alcohol | Type | N/A | 140 | 138 | 54 | 0.39 | 0.39 | 0.39 |
| Drug | Amount | N/A | 5 | 0 | 0 | 0.00 | 0.00 | 0.00 |
| Drug | Duration | N/A | 13 | 0 | 0 | 0.00 | 0.00 | 0.00 |
| Drug | Frequency | N/A | 21 | 0 | 0 | 0.00 | 0.00 | 0.00 |
| Drug | History | N/A | 71 | 0 | 0 | 0.00 | 0.00 | 0.00 |

| Drug | Method | N/A | 146 | 83 | 45 | 0.54 | 0.31 | 0.39 |
| Drug | StatusTime | current | 128 | 36 | 14 | 0.39 | 0.11 | 0.17 |
| Drug | StatusTime | none | 491 | 386 | 315 | 0.82 | 0.64 | 0.72 |
| Drug | StatusTime | past | 150 | 23 | 11 | 0.48 | 0.07 | 0.13 |
| Drug | Trigger | N/A | 769 | 745 | 637 | 0.86 | 0.83 | 0.84 |
| Drug | Type | N/A | 494 | 408 | 261 | 0.64 | 0.53 | 0.58 |
| Employment | Duration | N/A | 30 | 0 | 0 | 0.00 | 0.00 | 0.00 |
| Employment | History | N/A | 35 | 0 | 0 | 0.00 | 0.00 | 0.00 |
| Employment | StatusEmploy | employed | 223 | 161 | 122 | 0.76 | 0.55 | 0.64 |
| Employment | StatusEmploy | homemaker | 9 | 0 | 0 | 0.00 | 0.00 | 0.00 |
| Employment | StatusEmploy | on_disability | 102 | 55 | 40 | 0.73 | 0.39 | 0.51 |
| Employment | StatusEmploy | retired | 141 | 109 | 99 | 0.91 | 0.70 | 0.79 |
| Employment | StatusEmploy | student | 13 | 3 | 3 | 1.00 | 0.23 | 0.38 |
| Employment | StatusEmploy | unemployed | 275 | 214 | 138 | 0.64 | 0.50 | 0.56 |
| Employment | Trigger | N/A | 763 | 740 | 566 | 0.76 | 0.74 | 0.75 |
| Employment | Type | N/A | 559 | 172 | 59 | 0.34 | 0.11 | 0.16 |
| LivingStatus | Duration | N/A | 23 | 0 | 0 | 0.00 | 0.00 | 0.00 |
| LivingStatus | History | N/A | 10 | 0 | 0 | 0.00 | 0.00 | 0.00 |
| LivingStatus | StatusTime | current | 761 | 715 | 651 | 0.91 | 0.86 | 0.88 |
| LivingStatus | StatusTime | future | 1 | 0 | 0 | 0.00 | 0.00 | 0.00 |
| LivingStatus | StatusTime | none | 1 | 0 | 0 | 0.00 | 0.00 | 0.00 |
| LivingStatus | StatusTime | past | 49 | 1 | 0 | 0.00 | 0.00 | 0.00 |
| LivingStatus | Trigger | N/A | 812 | 773 | 690 | 0.89 | 0.85 | 0.87 |
| LivingStatus | TypeLiving | alone | 169 | 145 | 130 | 0.90 | 0.77 | 0.83 |
| LivingStatus | TypeLiving | homeless | 33 | 13 | 12 | 0.92 | 0.36 | 0.52 |
| LivingStatus | TypeLiving | with_family | 447 | 364 | 329 | 0.90 | 0.74 | 0.81 |
| LivingStatus | TypeLiving | with_others | 163 | 109 | 57 | 0.52 | 0.35 | 0.42 |
| Tobacco | Amount | N/A | 445 | 214 | 123 | 0.57 | 0.28 | 0.37 |
| Tobacco | Duration | N/A | 215 | 35 | 21 | 0.60 | 0.10 | 0.17 |
| Tobacco | Frequency | N/A | 141 | 107 | 60 | 0.56 | 0.43 | 0.48 |
| Tobacco | History | N/A | 266 | 115 | 64 | 0.56 | 0.24 | 0.34 |
| Tobacco | Method | N/A | 3 | 0 | 0 | 0.00 | 0.00 | 0.00 |
| Tobacco | StatusTime | current | 312 | 133 | 95 | 0.71 | 0.30 | 0.43 |
| Tobacco | StatusTime | none | 355 | 248 | 218 | 0.88 | 0.61 | 0.72 |
| Tobacco | StatusTime | past | 383 | 238 | 177 | 0.74 | 0.46 | 0.57 |
| Tobacco | Trigger | N/A | 1050 | 956 | 825 | 0.86 | 0.79 | 0.82 |
| Tobacco | Type | N/A | 96 | 59 | 25 | 0.42 | 0.26 | 0.32 |

**Table A5:** Precision, recall and F1 score obtained when adapting the SDoH information extraction model from **UW data with 400 samples from the MIMIC-III train set.** NT is the count of true (gold) labels, NP is the count of predicted labels, TP is the counted true positives, P is precision, R is recall and F1 is the F-1 measure, which is the harmonic mean of precision and recall.

| SDoH event | Argument | Subtype | NT | NP | TP | P | R | F1 |
|---|---|---|---|---|---|---|---|---|
| OVERALL | OVERALL | OVERALL | 13110 | 9424 | 7336 | 0.78 | 0.56 | 0.65 |
| Alcohol | Amount | N/A | 236 | 28 | 18 | 0.64 | 0.08 | 0.14 |
| Alcohol | Duration | N/A | 52 | 0 | 0 | 0.00 | 0.00 | 0.00 |
| Alcohol | Frequency | N/A | 233 | 103 | 47 | 0.46 | 0.20 | 0.28 |
| Alcohol | History | N/A | 116 | 0 | 0 | 0.00 | 0.00 | 0.00 |
| Alcohol | StatusTime | current | 460 | 273 | 176 | 0.64 | 0.38 | 0.48 |
| Alcohol | StatusTime | none | 391 | 316 | 253 | 0.80 | 0.65 | 0.72 |
| Alcohol | StatusTime | past | 229 | 68 | 36 | 0.53 | 0.16 | 0.24 |
| Alcohol | Trigger | N/A | 1080 | 1064 | 876 | 0.82 | 0.81 | 0.82 |
| Alcohol | Type | N/A | 140 | 107 | 46 | 0.43 | 0.33 | 0.37 |
| Drug | Amount | N/A | 5 | 0 | 0 | 0.00 | 0.00 | 0.00 |
| Drug | Duration | N/A | 13 | 0 | 0 | 0.00 | 0.00 | 0.00 |
| Drug | Frequency | N/A | 21 | 0 | 0 | 0.00 | 0.00 | 0.00 |
| Drug | History | N/A | 71 | 0 | 0 | 0.00 | 0.00 | 0.00 |
| Drug | Method | N/A | 146 | 101 | 56 | 0.55 | 0.38 | 0.45 |
| Drug | StatusTime | current | 128 | 26 | 12 | 0.46 | 0.09 | 0.16 |
| Drug | StatusTime | none | 491 | 396 | 319 | 0.81 | 0.65 | 0.72 |
| Drug | StatusTime | past | 150 | 28 | 12 | 0.43 | 0.08 | 0.13 |
| Drug | Trigger | N/A | 769 | 764 | 648 | 0.85 | 0.84 | 0.85 |
| Drug | Type | N/A | 494 | 424 | 260 | 0.61 | 0.53 | 0.57 |
| Employment | Duration | N/A | 30 | 0 | 0 | 0.00 | 0.00 | 0.00 |
| Employment | History | N/A | 35 | 1 | 0 | 0.00 | 0.00 | 0.00 |
| Employment | StatusEmploy | employed | 223 | 143 | 119 | 0.83 | 0.53 | 0.65 |
| Employment | StatusEmploy | homemaker | 9 | 0 | 0 | 0.00 | 0.00 | 0.00 |
| Employment | StatusEmploy | on_disability | 102 | 61 | 45 | 0.74 | 0.44 | 0.55 |
| Employment | StatusEmploy | retired | 141 | 115 | 105 | 0.91 | 0.74 | 0.82 |
| Employment | StatusEmploy | student | 13 | 2 | 2 | 1.00 | 0.15 | 0.27 |
| Employment | StatusEmploy | unemployed | 275 | 220 | 140 | 0.64 | 0.51 | 0.57 |
| Employment | Trigger | N/A | 763 | 682 | 535 | 0.78 | 0.70 | 0.74 |
| Employment | Type | N/A | 559 | 180 | 71 | 0.39 | 0.13 | 0.19 |
| LivingStatus | Duration | N/A | 23 | 0 | 0 | 0.00 | 0.00 | 0.00 |
| LivingStatus | History | N/A | 10 | 0 | 0 | 0.00 | 0.00 | 0.00 |
| LivingStatus | StatusTime | current | 761 | 752 | 666 | 0.89 | 0.88 | 0.88 |

| SDoH event | Argument | Subtype | NT | NP | TP | P | R | F1 |
|---|---|---|---|---|---|---|---|---|
| LivingStatus | StatusTime | future | 1 | 0 | 0 | 0.00 | 0.00 | 0.00 |
| LivingStatus | StatusTime | none | 1 | 0 | 0 | 0.00 | 0.00 | 0.00 |
| LivingStatus | StatusTime | past | 49 | 0 | 0 | 0.00 | 0.00 | 0.00 |
| LivingStatus | Trigger | N/A | 812 | 818 | 718 | 0.88 | 0.88 | 0.88 |
| LivingStatus | TypeLiving | alone | 169 | 149 | 137 | 0.92 | 0.81 | 0.86 |
| LivingStatus | TypeLiving | homeless | 33 | 18 | 16 | 0.89 | 0.48 | 0.63 |
| LivingStatus | TypeLiving | with_family | 447 | 386 | 344 | 0.89 | 0.77 | 0.83 |
| LivingStatus | TypeLiving | with_others | 163 | 109 | 51 | 0.47 | 0.31 | 0.38 |
| Tobacco | Amount | N/A | 445 | 210 | 115 | 0.55 | 0.26 | 0.35 |
| Tobacco | Duration | N/A | 215 | 25 | 14 | 0.56 | 0.07 | 0.12 |
| Tobacco | Frequency | N/A | 141 | 97 | 54 | 0.56 | 0.38 | 0.45 |
| Tobacco | History | N/A | 266 | 94 | 58 | 0.62 | 0.22 | 0.32 |
| Tobacco | Method | N/A | 3 | 0 | 0 | 0.00 | 0.00 | 0.00 |
| Tobacco | StatusTime | current | 312 | 123 | 88 | 0.72 | 0.28 | 0.40 |
| Tobacco | StatusTime | none | 355 | 278 | 236 | 0.85 | 0.66 | 0.75 |
| Tobacco | StatusTime | past | 383 | 250 | 195 | 0.78 | 0.51 | 0.62 |
| Tobacco | Trigger | N/A | 1050 | 978 | 852 | 0.87 | 0.81 | 0.84 |
| Tobacco | Type | N/A | 96 | 35 | 16 | 0.46 | 0.17 | 0.24 |

Table A6: Precision, recall and F1 score obtained when adapting the SDoH information extraction model from **UW data with 500 samples from the MIMIC-III train set.** NT is the count of true (gold) labels, NP is the count of predicted labels, TP is the counted true positives, P is precision, R is recall and F1 is the F-1 measure, which is the harmonic mean of precision and recall.

| SDoH event | Argument | Subtype | NT | NP | TP | P | R | F1 |
|---|---|---|---|---|---|---|---|---|
| OVERALL | OVERALL | OVERALL | 13110 | 9728 | 7455 | 0.77 | 0.57 | 0.65 |
| Alcohol | Amount | N/A | 236 | 43 | 25 | 0.58 | 0.11 | 0.18 |
| Alcohol | Duration | N/A | 52 | 0 | 0 | 0.00 | 0.00 | 0.00 |
| Alcohol | Frequency | N/A | 233 | 114 | 59 | 0.52 | 0.25 | 0.34 |
| Alcohol | History | N/A | 116 | 0 | 0 | 0.00 | 0.00 | 0.00 |
| Alcohol | StatusTime | current | 460 | 283 | 177 | 0.63 | 0.38 | 0.48 |
| Alcohol | StatusTime | none | 391 | 355 | 282 | 0.79 | 0.72 | 0.76 |
| Alcohol | StatusTime | past | 229 | 43 | 22 | 0.51 | 0.10 | 0.16 |
| Alcohol | Trigger | N/A | 1080 | 1128 | 911 | 0.81 | 0.84 | 0.83 |
| Alcohol | Type | N/A | 140 | 112 | 45 | 0.40 | 0.32 | 0.36 |
| Drug | Amount | N/A | 5 | 0 | 0 | 0.00 | 0.00 | 0.00 |
| Drug | Duration | N/A | 13 | 0 | 0 | 0.00 | 0.00 | 0.00 |
| Drug | Frequency | N/A | 21 | 0 | 0 | 0.00 | 0.00 | 0.00 |
| Drug | History | N/A | 71 | 0 | 0 | 0.00 | 0.00 | 0.00 |
| Drug | Method | N/A | 146 | 81 | 40 | 0.49 | 0.27 | 0.35 |
| Drug | StatusTime | current | 128 | 43 | 16 | 0.37 | 0.13 | 0.19 |
| Drug | StatusTime | none | 491 | 393 | 304 | 0.77 | 0.62 | 0.69 |
| Drug | StatusTime | past | 150 | 29 | 11 | 0.38 | 0.07 | 0.12 |
| Drug | Trigger | N/A | 769 | 805 | 661 | 0.82 | 0.86 | 0.84 |
| Drug | Type | N/A | 494 | 433 | 254 | 0.59 | 0.51 | 0.55 |
| Employment | Duration | N/A | 30 | 0 | 0 | 0.00 | 0.00 | 0.00 |
| Employment | History | N/A | 35 | 0 | 0 | 0.00 | 0.00 | 0.00 |
| Employment | StatusEmploy | employed | 223 | 168 | 123 | 0.73 | 0.55 | 0.63 |
| Employment | StatusEmploy | homemaker | 9 | 0 | 0 | 0.00 | 0.00 | 0.00 |
| Employment | StatusEmploy | on_disability | 102 | 58 | 40 | 0.69 | 0.39 | 0.50 |
| Employment | StatusEmploy | retired | 141 | 128 | 114 | 0.89 | 0.81 | 0.85 |
| Employment | StatusEmploy | student | 13 | 6 | 4 | 0.67 | 0.31 | 0.42 |
| Employment | StatusEmploy | unemployed | 275 | 242 | 148 | 0.61 | 0.54 | 0.57 |
| Employment | Trigger | N/A | 763 | 732 | 559 | 0.76 | 0.73 | 0.75 |
| Employment | Type | N/A | 559 | 140 | 61 | 0.44 | 0.11 | 0.17 |
| LivingStatus | Duration | N/A | 23 | 0 | 0 | 0.00 | 0.00 | 0.00 |
| LivingStatus | History | N/A | 10 | 0 | 0 | 0.00 | 0.00 | 0.00 |
| LivingStatus | StatusTime | current | 761 | 716 | 641 | 0.90 | 0.84 | 0.87 |
| LivingStatus | StatusTime | future | 1 | 0 | 0 | 0.00 | 0.00 | 0.00 |
| LivingStatus | StatusTime | none | 1 | 0 | 0 | 0.00 | 0.00 | 0.00 |
| LivingStatus | StatusTime | past | 49 | 0 | 0 | 0.00 | 0.00 | 0.00 |
| LivingStatus | Trigger | N/A | 812 | 784 | 703 | 0.90 | 0.87 | 0.88 |
| LivingStatus | TypeLiving | alone | 169 | 150 | 131 | 0.87 | 0.78 | 0.82 |
| LivingStatus | TypeLiving | homeless | 33 | 20 | 17 | 0.85 | 0.52 | 0.64 |
| LivingStatus | TypeLiving | with_family | 447 | 382 | 348 | 0.91 | 0.78 | 0.84 |
| LivingStatus | TypeLiving | with_others | 163 | 80 | 42 | 0.53 | 0.26 | 0.35 |
| Tobacco | Amount | N/A | 445 | 197 | 112 | 0.57 | 0.25 | 0.35 |
| Tobacco | Duration | N/A | 215 | 40 | 20 | 0.50 | 0.09 | 0.16 |
| Tobacco | Frequency | N/A | 141 | 84 | 48 | 0.57 | 0.34 | 0.43 |
| Tobacco | History | N/A | 266 | 141 | 76 | 0.54 | 0.29 | 0.37 |
| Tobacco | Method | N/A | 3 | 0 | 0 | 0.00 | 0.00 | 0.00 |
| Tobacco | StatusTime | current | 312 | 158 | 94 | 0.59 | 0.30 | 0.40 |
| Tobacco | StatusTime | none | 355 | 266 | 241 | 0.91 | 0.68 | 0.78 |
| Tobacco | StatusTime | past | 383 | 276 | 210 | 0.76 | 0.55 | 0.64 |
| Tobacco | Trigger | N/A | 1050 | 1025 | 886 | 0.86 | 0.84 | 0.85 |
| Tobacco | Type | N/A | 96 | 73 | 30 | 0.41 | 0.31 | 0.36 |

**Table A7:** Precision, recall and F1 score obtained when training the SDoH information extraction model **from scratch on 100 samples from the MIMIC-III train set**. NT is the count of true (gold) labels, NP is the count of predicted labels, TP is the counted true positives, P is precision, R is recall and F1 is the F-1 measure, which is the harmonic mean of precision and recall.

| SDoH event | Argument | Subtype | NT | NP | TP | P | R | F1 |
| --- | --- | --- | --- | --- | --- | --- | --- | --- |
| OVERALL | OVERALL | OVERALL | 13110 | 8943 | 6334 | 0.71 | 0.48 | 0.57 |
| Alcohol | Amount | N/A | 236 | 46 | 11 | 0.24 | 0.05 | 0.08 |
| Alcohol | Duration | N/A | 52 | 3 | 0 | 0.00 | 0.00 | 0.00 |
| Alcohol | Frequency | N/A | 233 | 68 | 29 | 0.43 | 0.12 | 0.19 |
| Alcohol | History | N/A | 116 | 10 | 0 | 0.00 | 0.00 | 0.00 |
| Alcohol | StatusTime | current | 460 | 159 | 92 | 0.58 | 0.20 | 0.30 |
| Alcohol | StatusTime | none | 391 | 211 | 167 | 0.79 | 0.43 | 0.55 |
| Alcohol | StatusTime | past | 229 | 20 | 8 | 0.40 | 0.03 | 0.06 |
| Alcohol | Trigger | N/A | 1080 | 1073 | 866 | 0.81 | 0.80 | 0.80 |
| Alcohol | Type | N/A | 140 | 93 | 37 | 0.40 | 0.26 | 0.32 |
| Drug | Amount | N/A | 5 | 0 | 0 | 0.00 | 0.00 | 0.00 |
| Drug | Duration | N/A | 13 | 0 | 0 | 0.00 | 0.00 | 0.00 |
| Drug | Frequency | N/A | 21 | 0 | 0 | 0.00 | 0.00 | 0.00 |
| Drug | History | N/A | 71 | 1 | 0 | 0.00 | 0.00 | 0.00 |
| Drug | Method | N/A | 146 | 183 | 74 | 0.40 | 0.51 | 0.45 |
| Drug | StatusTime | current | 128 | 24 | 7 | 0.29 | 0.05 | 0.09 |
| Drug | StatusTime | none | 491 | 269 | 196 | 0.73 | 0.40 | 0.52 |
| Drug | StatusTime | past | 150 | 4 | 2 | 0.50 | 0.01 | 0.03 |
| Drug | Trigger | N/A | 769 | 719 | 548 | 0.76 | 0.71 | 0.74 |
| Drug | Type | N/A | 494 | 417 | 207 | 0.50 | 0.42 | 0.45 |
| Employment | Duration | N/A | 30 | 0 | 0 | 0.00 | 0.00 | 0.00 |
| Employment | History | N/A | 35 | 0 | 0 | 0.00 | 0.00 | 0.00 |
| Employment | StatusEmploy | employed | 223 | 137 | 112 | 0.82 | 0.50 | 0.62 |
| Employment | StatusEmploy | homemaker | 9 | 3 | 1 | 0.33 | 0.11 | 0.17 |
| Employment | StatusEmploy | on_disability | 102 | 35 | 26 | 0.74 | 0.25 | 0.38 |
| Employment | StatusEmploy | retired | 141 | 99 | 81 | 0.82 | 0.57 | 0.68 |
| Employment | StatusEmploy | student | 13 | 11 | 3 | 0.27 | 0.23 | 0.25 |
| Employment | StatusEmploy | unemployed | 275 | 162 | 99 | 0.61 | 0.36 | 0.45 |
| Employment | Trigger | N/A | 763 | 682 | 495 | 0.73 | 0.65 | 0.69 |
| Employment | Type | N/A | 559 | 258 | 66 | 0.26 | 0.12 | 0.16 |
| LivingStatus | Duration | N/A | 23 | 0 | 0 | 0.00 | 0.00 | 0.00 |
| LivingStatus | History | N/A | 10 | 0 | 0 | 0.00 | 0.00 | 0.00 |
| LivingStatus | StatusTime | current | 761 | 775 | 655 | 0.85 | 0.86 | 0.85 |

| SDoH event | Argument | Subtype | NT | NP | TP | P | R | F1 |
|---|---|---|---|---|---|---|---|---|
| LivingStatus | StatusTime | future | 1 | 0 | 0 | 0.00 | 0.00 | 0.00 |
| LivingStatus | StatusTime | none | 1 | 0 | 0 | 0.00 | 0.00 | 0.00 |
| LivingStatus | StatusTime | past | 49 | 0 | 0 | 0.00 | 0.00 | 0.00 |
| LivingStatus | Trigger | N/A | 812 | 819 | 709 | 0.87 | 0.87 | 0.87 |
| LivingStatus | TypeLiving | alone | 169 | 146 | 125 | 0.86 | 0.74 | 0.79 |
| LivingStatus | TypeLiving | homeless | 33 | 9 | 8 | 0.89 | 0.24 | 0.38 |
| LivingStatus | TypeLiving | with_family | 447 | 379 | 331 | 0.87 | 0.74 | 0.80 |
| LivingStatus | TypeLiving | with_others | 163 | 98 | 38 | 0.39 | 0.23 | 0.29 |
| Tobacco | Amount | N/A | 445 | 236 | 90 | 0.38 | 0.20 | 0.26 |
| Tobacco | Duration | N/A | 215 | 18 | 11 | 0.61 | 0.05 | 0.09 |
| Tobacco | Frequency | N/A | 141 | 47 | 6 | 0.13 | 0.04 | 0.06 |
| Tobacco | History | N/A | 266 | 81 | 29 | 0.36 | 0.11 | 0.17 |
| Tobacco | Method | N/A | 3 | 0 | 0 | 0.00 | 0.00 | 0.00 |
| Tobacco | StatusTime | current | 312 | 108 | 61 | 0.56 | 0.20 | 0.29 |
| Tobacco | StatusTime | none | 355 | 226 | 153 | 0.68 | 0.43 | 0.53 |
| Tobacco | StatusTime | past | 383 | 242 | 161 | 0.67 | 0.42 | 0.52 |
| Tobacco | Trigger | N/A | 1050 | 1027 | 810 | 0.79 | 0.77 | 0.78 |
| Tobacco | Type | N/A | 96 | 45 | 20 | 0.44 | 0.21 | 0.28 |

Table A8: Precision, recall and F1 score obtained when training the SDoH information extraction model **from scratch on 200 samples from the MIMIC-III train set**. NT is the count of true (gold) labels, NP is the count of predicted labels, TP is the counted true positives, P is precision, R is recall and F1 is the F-1 measure, which is the harmonic mean of precision and recall.

| SDoH event | Argument | Subtype | NT | NP | TP | P | R | F1 |
|---|---|---|---|---|---|---|---|---|
| OVERALL | OVERALL | OVERALL | 13110 | 9260 | 6767 | 0.73 | 0.52 | 0.61 |
| Alcohol | Amount | N/A | 236 | 66 | 19 | 0.29 | 0.08 | 0.13 |
| Alcohol | Duration | N/A | 52 | 2 | 0 | 0.00 | 0.00 | 0.00 |
| Alcohol | Frequency | N/A | 233 | 150 | 55 | 0.37 | 0.24 | 0.29 |
| Alcohol | History | N/A | 116 | 2 | 0 | 0.00 | 0.00 | 0.00 |
| Alcohol | StatusTime | current | 460 | 240 | 147 | 0.61 | 0.32 | 0.42 |
| Alcohol | StatusTime | none | 391 | 234 | 183 | 0.78 | 0.47 | 0.59 |
| Alcohol | StatusTime | past | 229 | 46 | 15 | 0.33 | 0.07 | 0.11 |
| Alcohol | Trigger | N/A | 1080 | 1048 | 868 | 0.83 | 0.80 | 0.82 |
| Alcohol | Type | N/A | 140 | 96 | 48 | 0.50 | 0.34 | 0.41 |
| Drug | Amount | N/A | 5 | 0 | 0 | 0.00 | 0.00 | 0.00 |
| Drug | Duration | N/A | 13 | 0 | 0 | 0.00 | 0.00 | 0.00 |
| Drug | Frequency | N/A | 21 | 0 | 0 | 0.00 | 0.00 | 0.00 |
| Drug | History | N/A | 71 | 0 | 0 | 0.00 | 0.00 | 0.00 |
| Drug | Method | N/A | 146 | 141 | 68 | 0.48 | 0.47 | 0.47 |
| Drug | StatusTime | current | 128 | 18 | 3 | 0.17 | 0.02 | 0.04 |
| Drug | StatusTime | none | 491 | 316 | 241 | 0.76 | 0.49 | 0.60 |
| Drug | StatusTime | past | 150 | 20 | 4 | 0.20 | 0.03 | 0.05 |
| Drug | Trigger | N/A | 769 | 735 | 599 | 0.81 | 0.78 | 0.80 |
| Drug | Type | N/A | 494 | 417 | 240 | 0.58 | 0.49 | 0.53 |
| Employment | Duration | N/A | 30 | 0 | 0 | 0.00 | 0.00 | 0.00 |
| Employment | History | N/A | 35 | 0 | 0 | 0.00 | 0.00 | 0.00 |
| Employment | StatusEmploy | employed | 223 | 117 | 98 | 0.84 | 0.44 | 0.58 |
| Employment | StatusEmploy | homemaker | 9 | 0 | 0 | 0.00 | 0.00 | 0.00 |
| Employment | StatusEmploy | on_disability | 102 | 60 | 42 | 0.70 | 0.41 | 0.52 |
| Employment | StatusEmploy | retired | 141 | 134 | 108 | 0.81 | 0.77 | 0.79 |
| Employment | StatusEmploy | student | 13 | 10 | 2 | 0.20 | 0.15 | 0.17 |
| Employment | StatusEmploy | unemployed | 275 | 163 | 98 | 0.60 | 0.36 | 0.45 |
| Employment | Trigger | N/A | 763 | 682 | 512 | 0.75 | 0.67 | 0.71 |
| Employment | Type | N/A | 559 | 285 | 101 | 0.35 | 0.18 | 0.24 |
| LivingStatus | Duration | N/A | 23 | 2 | 0 | 0.00 | 0.00 | 0.00 |
| LivingStatus | History | N/A | 10 | 0 | 0 | 0.00 | 0.00 | 0.00 |
| LivingStatus | StatusTime | current | 761 | 737 | 635 | 0.86 | 0.83 | 0.85 |
| LivingStatus | StatusTime | future | 1 | 0 | 0 | 0.00 | 0.00 | 0.00 |
| LivingStatus | StatusTime | none | 1 | 0 | 0 | 0.00 | 0.00 | 0.00 |
| LivingStatus | StatusTime | past | 49 | 2 | 2 | 1.00 | 0.04 | 0.08 |
| LivingStatus | Trigger | N/A | 812 | 761 | 658 | 0.86 | 0.81 | 0.84 |
| LivingStatus | TypeLiving | alone | 169 | 134 | 117 | 0.87 | 0.69 | 0.77 |
| LivingStatus | TypeLiving | homeless | 33 | 10 | 8 | 0.80 | 0.24 | 0.37 |
| LivingStatus | TypeLiving | with_family | 447 | 354 | 315 | 0.89 | 0.70 | 0.79 |
| LivingStatus | TypeLiving | with_others | 163 | 74 | 39 | 0.53 | 0.24 | 0.33 |
| Tobacco | Amount | N/A | 445 | 217 | 107 | 0.49 | 0.24 | 0.32 |
| Tobacco | Duration | N/A | 215 | 112 | 39 | 0.35 | 0.18 | 0.24 |
| Tobacco | Frequency | N/A | 141 | 94 | 46 | 0.49 | 0.33 | 0.39 |
| Tobacco | History | N/A | 266 | 113 | 46 | 0.41 | 0.17 | 0.24 |
| Tobacco | Method | N/A | 3 | 0 | 0 | 0.00 | 0.00 | 0.00 |
| Tobacco | StatusTime | current | 312 | 158 | 90 | 0.57 | 0.29 | 0.38 |
| Tobacco | StatusTime | none | 355 | 198 | 172 | 0.87 | 0.48 | 0.62 |
| Tobacco | StatusTime | past | 383 | 229 | 164 | 0.72 | 0.43 | 0.54 |
| Tobacco | Trigger | N/A | 1050 | 1043 | 857 | 0.82 | 0.82 | 0.82 |
| Tobacco | Type | N/A | 96 | 40 | 21 | 0.53 | 0.22 | 0.31 |

**Table A9:** Precision, recall and F1 score obtained when training the SDoH information extraction model **from scratch on 300 samples from the MIMIC-III train set**. NT is the count of true (gold) labels, NP is the count of predicted labels, TP is the counted true positives, P is precision, R is recall and F1 is the F-1 measure, which is the harmonic mean of precision and recall.

| SDoH event | Argument | Subtype | NT | NP | TP | P | R | F1 |
|---|---|---|---|---|---|---|---|---|
| OVERALL | OVERALL | OVERALL | 13110 | 9965 | 7349 | 0.74 | 0.56 | 0.64 |
| Alcohol | Amount | N/A | 236 | 94 | 30 | 0.32 | 0.13 | 0.18 |
| Alcohol | Duration | N/A | 52 | 0 | 0 | 0.00 | 0.00 | 0.00 |
| Alcohol | Frequency | N/A | 233 | 168 | 73 | 0.43 | 0.31 | 0.36 |
| Alcohol | History | N/A | 116 | 0 | 0 | 0.00 | 0.00 | 0.00 |
| Alcohol | StatusTime | current | 460 | 298 | 176 | 0.59 | 0.38 | 0.46 |
| Alcohol | StatusTime | none | 391 | 307 | 235 | 0.77 | 0.60 | 0.67 |
| Alcohol | StatusTime | past | 229 | 34 | 14 | 0.41 | 0.06 | 0.11 |
| Alcohol | Trigger | N/A | 1080 | 1130 | 916 | 0.81 | 0.85 | 0.83 |
| Alcohol | Type | N/A | 140 | 125 | 52 | 0.42 | 0.37 | 0.39 |
| Drug | Amount | N/A | 5 | 0 | 0 | 0.00 | 0.00 | 0.00 |
| Drug | Duration | N/A | 13 | 0 | 0 | 0.00 | 0.00 | 0.00 |
| Drug | Frequency | N/A | 21 | 0 | 0 | 0.00 | 0.00 | 0.00 |
| Drug | History | N/A | 71 | 0 | 0 | 0.00 | 0.00 | 0.00 |
| Drug | Method | N/A | 146 | 68 | 34 | 0.50 | 0.23 | 0.32 |
| Drug | StatusTime | current | 128 | 29 | 10 | 0.34 | 0.08 | 0.13 |
| Drug | StatusTime | none | 491 | 457 | 348 | 0.76 | 0.71 | 0.73 |
| Drug | StatusTime | past | 150 | 18 | 10 | 0.56 | 0.07 | 0.12 |
| Drug | Trigger | N/A | 769 | 800 | 637 | 0.80 | 0.83 | 0.81 |
| Drug | Type | N/A | 494 | 435 | 239 | 0.55 | 0.48 | 0.51 |
| Employment | Duration | N/A | 30 | 0 | 0 | 0.00 | 0.00 | 0.00 |
| Employment | History | N/A | 35 | 0 | 0 | 0.00 | 0.00 | 0.00 |
| Employment | StatusEmploy | employed | 223 | 130 | 107 | 0.82 | 0.48 | 0.61 |
| Employment | StatusEmploy | homemaker | 9 | 3 | 1 | 0.33 | 0.11 | 0.17 |
| Employment | StatusEmploy | on_disability | 102 | 72 | 55 | 0.76 | 0.54 | 0.63 |
| Employment | StatusEmploy | retired | 141 | 110 | 94 | 0.85 | 0.67 | 0.75 |
| Employment | StatusEmploy | student | 13 | 11 | 2 | 0.18 | 0.15 | 0.17 |
| Employment | StatusEmploy | unemployed | 275 | 204 | 125 | 0.61 | 0.45 | 0.52 |
| Employment | Trigger | N/A | 763 | 722 | 529 | 0.73 | 0.69 | 0.71 |
| Employment | Type | N/A | 559 | 288 | 104 | 0.36 | 0.19 | 0.25 |
| LivingStatus | Duration | N/A | 23 | 2 | 0 | 0.00 | 0.00 | 0.00 |
| LivingStatus | History | N/A | 10 | 0 | 0 | 0.00 | 0.00 | 0.00 |
| LivingStatus | StatusTime | current | 761 | 743 | 647 | 0.87 | 0.85 | 0.86 |

| SDoH event | Argument | Subtype | NT | NP | TP | P | R | F1 |
|---|---|---|---|---|---|---|---|---|
| LivingStatus | StatusTime | future | 1 | 0 | 0 | 0.00 | 0.00 | 0.00 |
| LivingStatus | StatusTime | none | 1 | 0 | 0 | 0.00 | 0.00 | 0.00 |
| LivingStatus | StatusTime | past | 49 | 8 | 5 | 0.63 | 0.10 | 0.18 |
| LivingStatus | Trigger | N/A | 812 | 817 | 700 | 0.86 | 0.86 | 0.86 |
| LivingStatus | TypeLiving | alone | 169 | 154 | 132 | 0.86 | 0.78 | 0.82 |
| LivingStatus | TypeLiving | homeless | 33 | 8 | 8 | 1.00 | 0.24 | 0.39 |
| LivingStatus | TypeLiving | with_family | 447 | 416 | 353 | 0.85 | 0.79 | 0.82 |
| LivingStatus | TypeLiving | with_others | 163 | 62 | 42 | 0.68 | 0.26 | 0.37 |
| Tobacco | Amount | N/A | 445 | 247 | 118 | 0.48 | 0.27 | 0.34 |
| Tobacco | Duration | N/A | 215 | 113 | 49 | 0.43 | 0.23 | 0.30 |
| Tobacco | Frequency | N/A | 141 | 101 | 64 | 0.63 | 0.45 | 0.53 |
| Tobacco | History | N/A | 266 | 97 | 49 | 0.51 | 0.18 | 0.27 |
| Tobacco | Method | N/A | 3 | 0 | 0 | 0.00 | 0.00 | 0.00 |
| Tobacco | StatusTime | current | 312 | 153 | 100 | 0.65 | 0.32 | 0.43 |
| Tobacco | StatusTime | none | 355 | 265 | 221 | 0.83 | 0.62 | 0.71 |
| Tobacco | StatusTime | past | 383 | 244 | 188 | 0.77 | 0.49 | 0.60 |
| Tobacco | Trigger | N/A | 1050 | 990 | 863 | 0.87 | 0.82 | 0.85 |
| Tobacco | Type | N/A | 96 | 42 | 19 | 0.45 | 0.20 | 0.28 |

Table A10: Precision, recall and F1 score obtained when training the SDoH information extraction model **from scratch on 400 samples from the MIMIC-III train set**. NT is the count of true (gold) labels, NP is the count of predicted labels, TP is the counted true positives, P is precision, R is recall and F1 is the F-1 measure, which is the harmonic mean of precision and recall.

| SDoH event | Argument | Subtype | NT | NP | TP | P | R | F1 |
|---|---|---|---|---|---|---|---|---|
| OVERALL | OVERALL | OVERALL | 13110 | 9804 | 7209 | 0.74 | 0.55 | 0.63 |
| Alcohol | Amount | N/A | 236 | 103 | 36 | 0.35 | 0.15 | 0.21 |
| Alcohol | Duration | N/A | 52 | 0 | 0 | 0.00 | 0.00 | 0.00 |
| Alcohol | Frequency | N/A | 233 | 146 | 63 | 0.43 | 0.27 | 0.33 |
| Alcohol | History | N/A | 116 | 0 | 0 | 0.00 | 0.00 | 0.00 |
| Alcohol | StatusTime | current | 460 | 324 | 199 | 0.61 | 0.43 | 0.51 |
| Alcohol | StatusTime | none | 391 | 288 | 219 | 0.76 | 0.56 | 0.65 |
| Alcohol | StatusTime | past | 229 | 55 | 26 | 0.47 | 0.11 | 0.18 |
| Alcohol | Trigger | N/A | 1080 | 1098 | 885 | 0.81 | 0.82 | 0.81 |
| Alcohol | Type | N/A | 140 | 97 | 38 | 0.39 | 0.27 | 0.32 |
| Drug | Amount | N/A | 5 | 0 | 0 | 0.00 | 0.00 | 0.00 |
| Drug | Duration | N/A | 13 | 0 | 0 | 0.00 | 0.00 | 0.00 |
| Drug | Frequency | N/A | 21 | 0 | 0 | 0.00 | 0.00 | 0.00 |
| Drug | History | N/A | 71 | 0 | 0 | 0.00 | 0.00 | 0.00 |
| Drug | Method | N/A | 146 | 61 | 33 | 0.54 | 0.23 | 0.32 |
| Drug | StatusTime | current | 128 | 13 | 7 | 0.54 | 0.05 | 0.10 |
| Drug | StatusTime | none | 491 | 274 | 222 | 0.81 | 0.45 | 0.58 |
| Drug | StatusTime | past | 150 | 27 | 10 | 0.37 | 0.07 | 0.11 |
| Drug | Trigger | N/A | 769 | 738 | 596 | 0.81 | 0.78 | 0.79 |
| Drug | Type | N/A | 494 | 412 | 225 | 0.55 | 0.46 | 0.50 |
| Employment | Duration | N/A | 30 | 0 | 0 | 0.00 | 0.00 | 0.00 |
| Employment | History | N/A | 35 | 0 | 0 | 0.00 | 0.00 | 0.00 |
| Employment | StatusEmploy | employed | 223 | 164 | 125 | 0.76 | 0.56 | 0.65 |
| Employment | StatusEmploy | homemaker | 9 | 0 | 0 | 0.00 | 0.00 | 0.00 |
| Employment | StatusEmploy | on_disability | 102 | 75 | 58 | 0.77 | 0.57 | 0.66 |
| Employment | StatusEmploy | retired | 141 | 96 | 89 | 0.93 | 0.63 | 0.75 |
| Employment | StatusEmploy | student | 13 | 4 | 3 | 0.75 | 0.23 | 0.35 |
| Employment | StatusEmploy | unemployed | 275 | 189 | 113 | 0.60 | 0.41 | 0.49 |
| Employment | Trigger | N/A | 763 | 657 | 515 | 0.78 | 0.67 | 0.73 |
| Employment | Type | N/A | 559 | 263 | 105 | 0.40 | 0.19 | 0.26 |
| LivingStatus | Duration | N/A | 23 | 0 | 0 | 0.00 | 0.00 | 0.00 |
| LivingStatus | History | N/A | 10 | 0 | 0 | 0.00 | 0.00 | 0.00 |
| LivingStatus | StatusTime | current | 761 | 696 | 625 | 0.90 | 0.82 | 0.86 |
| LivingStatus | StatusTime | future | 1 | 0 | 0 | 0.00 | 0.00 | 0.00 |
| LivingStatus | StatusTime | none | 1 | 0 | 0 | 0.00 | 0.00 | 0.00 |
| LivingStatus | StatusTime | past | 49 | 5 | 2 | 0.40 | 0.04 | 0.07 |
| LivingStatus | Trigger | N/A | 812 | 747 | 668 | 0.89 | 0.82 | 0.86 |
| LivingStatus | TypeLiving | alone | 169 | 143 | 124 | 0.87 | 0.73 | 0.79 |
| LivingStatus | TypeLiving | homeless | 33 | 16 | 15 | 0.94 | 0.45 | 0.61 |
| LivingStatus | TypeLiving | with_family | 447 | 388 | 343 | 0.88 | 0.77 | 0.82 |
| LivingStatus | TypeLiving | with_others | 163 | 68 | 49 | 0.72 | 0.30 | 0.42 |
| Tobacco | Amount | N/A | 445 | 291 | 135 | 0.46 | 0.30 | 0.37 |
| Tobacco | Duration | N/A | 215 | 114 | 65 | 0.57 | 0.30 | 0.40 |
| Tobacco | Frequency | N/A | 141 | 125 | 70 | 0.56 | 0.50 | 0.53 |
| Tobacco | History | N/A | 266 | 126 | 54 | 0.43 | 0.20 | 0.28 |
| Tobacco | Method | N/A | 3 | 0 | 0 | 0.00 | 0.00 | 0.00 |
| Tobacco | StatusTime | current | 312 | 175 | 99 | 0.57 | 0.32 | 0.41 |
| Tobacco | StatusTime | none | 355 | 232 | 202 | 0.87 | 0.57 | 0.69 |
| Tobacco | StatusTime | past | 383 | 382 | 237 | 0.62 | 0.62 | 0.62 |
| Tobacco | Trigger | N/A | 1050 | 1129 | 914 | 0.81 | 0.87 | 0.84 |
| Tobacco | Type | N/A | 96 | 83 | 40 | 0.48 | 0.42 | 0.45 |

Table A11: Precision, recall and F1 score obtained when training the SDoH information extraction model **from scratch on 500 samples from the MIMIC-III train set**. NT is the count of true (gold) labels, NP is the count of predicted labels, TP is the counted true positives, P is precision, R is recall and F1 is the F-1 measure, which is the harmonic mean of precision and recall.

| SDoH event | Argument | Subtype | NT | NP | TP | P | R | F1 |
|---|---|---|---|---|---|---|---|---|
| OVERALL | OVERALL | OVERALL | 13110 | 9623 | 7326 | 0.76 | 0.56 | 0.64 |
| Alcohol | Amount | N/A | 236 | 78 | 36 | 0.46 | 0.15 | 0.23 |
| Alcohol | Duration | N/A | 52 | 0 | 0 | 0.00 | 0.00 | 0.00 |
| Alcohol | Frequency | N/A | 233 | 167 | 72 | 0.43 | 0.31 | 0.36 |
| Alcohol | History | N/A | 116 | 0 | 0 | 0.00 | 0.00 | 0.00 |
| Alcohol | StatusTime | current | 460 | 328 | 218 | 0.66 | 0.47 | 0.55 |
| Alcohol | StatusTime | none | 391 | 339 | 267 | 0.79 | 0.68 | 0.73 |
| Alcohol | StatusTime | past | 229 | 40 | 18 | 0.45 | 0.08 | 0.13 |
| Alcohol | Trigger | N/A | 1080 | 1028 | 872 | 0.85 | 0.81 | 0.83 |
| Alcohol | Type | N/A | 140 | 105 | 45 | 0.43 | 0.32 | 0.37 |
| Drug | Amount | N/A | 5 | 0 | 0 | 0.00 | 0.00 | 0.00 |
| Drug | Duration | N/A | 13 | 1 | 0 | 0.00 | 0.00 | 0.00 |
| Drug | Frequency | N/A | 21 | 0 | 0 | 0.00 | 0.00 | 0.00 |
| Drug | History | N/A | 71 | 0 | 0 | 0.00 | 0.00 | 0.00 |
| Drug | Method | N/A | 146 | 90 | 46 | 0.51 | 0.32 | 0.39 |
| Drug | StatusTime | N/A | 0 | 1 | 0 | 0.00 | 0.00 | 0.00 |
| Drug | StatusTime | current | 128 | 9 | 2 | 0.22 | 0.02 | 0.03 |
| Drug | StatusTime | none | 491 | 356 | 287 | 0.81 | 0.58 | 0.68 |
| Drug | StatusTime | past | 150 | 15 | 7 | 0.47 | 0.05 | 0.08 |
| Drug | Trigger | N/A | 769 | 718 | 599 | 0.83 | 0.78 | 0.81 |
| Drug | Type | N/A | 494 | 448 | 248 | 0.55 | 0.50 | 0.53 |
| Employment | Duration | N/A | 30 | 0 | 0 | 0.00 | 0.00 | 0.00 |
| Employment | History | N/A | 35 | 0 | 0 | 0.00 | 0.00 | 0.00 |
| Employment | StatusEmploy | employed | 223 | 145 | 115 | 0.79 | 0.52 | 0.63 |
| Employment | StatusEmploy | homemaker | 9 | 0 | 0 | 0.00 | 0.00 | 0.00 |
| Employment | StatusEmploy | on_disability | 102 | 73 | 55 | 0.75 | 0.54 | 0.63 |
| Employment | StatusEmploy | retired | 141 | 119 | 106 | 0.89 | 0.75 | 0.82 |
| Employment | StatusEmploy | student | 13 | 7 | 0 | 0.00 | 0.00 | 0.00 |
| Employment | StatusEmploy | unemployed | 275 | 213 | 138 | 0.65 | 0.50 | 0.57 |
| Employment | Trigger | N/A | 763 | 721 | 549 | 0.76 | 0.72 | 0.74 |
| Employment | Type | N/A | 559 | 309 | 122 | 0.39 | 0.22 | 0.28 |
| LivingStatus | Duration | N/A | 23 | 0 | 0 | 0.00 | 0.00 | 0.00 |
| LivingStatus | History | N/A | 10 | 0 | 0 | 0.00 | 0.00 | 0.00 |
| LivingStatus | StatusTime | current | 761 | 722 | 642 | 0.89 | 0.84 | 0.87 |
| LivingStatus | StatusTime | future | 1 | 0 | 0 | 0.00 | 0.00 | 0.00 |
| LivingStatus | StatusTime | none | 1 | 0 | 0 | 0.00 | 0.00 | 0.00 |
| LivingStatus | StatusTime | past | 49 | 10 | 8 | 0.80 | 0.16 | 0.27 |
| LivingStatus | Trigger | N/A | 812 | 794 | 701 | 0.88 | 0.86 | 0.87 |
| LivingStatus | TypeLiving | alone | 169 | 118 | 103 | 0.87 | 0.61 | 0.72 |
| LivingStatus | TypeLiving | homeless | 33 | 18 | 16 | 0.89 | 0.48 | 0.63 |
| LivingStatus | TypeLiving | with_family | 447 | 358 | 319 | 0.89 | 0.71 | 0.79 |
| LivingStatus | TypeLiving | with_others | 163 | 102 | 63 | 0.62 | 0.39 | 0.48 |
| Tobacco | Amount | N/A | 445 | 197 | 104 | 0.53 | 0.23 | 0.32 |
| Tobacco | Duration | N/A | 215 | 105 | 56 | 0.53 | 0.26 | 0.35 |
| Tobacco | Frequency | N/A | 141 | 98 | 53 | 0.54 | 0.38 | 0.44 |
| Tobacco | History | N/A | 266 | 110 | 63 | 0.57 | 0.24 | 0.34 |
| Tobacco | Method | N/A | 3 | 0 | 0 | 0.00 | 0.00 | 0.00 |
| Tobacco | StatusTime | current | 312 | 89 | 69 | 0.78 | 0.22 | 0.34 |
| Tobacco | StatusTime | none | 355 | 267 | 235 | 0.88 | 0.66 | 0.76 |
| Tobacco | StatusTime | past | 383 | 261 | 199 | 0.76 | 0.52 | 0.62 |
| Tobacco | Trigger | N/A | 1050 | 989 | 859 | 0.87 | 0.82 | 0.84 |
| Tobacco | Type | N/A | 96 | 75 | 34 | 0.45 | 0.35 | 0.40 |